\documentclass[11pt]{article}
\usepackage{appendix}
\usepackage[figuresright]{rotating}
\usepackage{amsmath,amssymb,amscd,amsthm}
\usepackage{graphicx}
\usepackage{dcolumn}
\usepackage{bm}
\usepackage{ifthen}
\usepackage{rays_defs_18}

\usepackage{hyperref}
\hypersetup{
    bookmarks=true,         
    unicode=false,          
    pdftoolbar=true,        
    pdfmenubar=true,        
    pdffitwindow=false,     
    pdfstartview={FitH},    
    pdfauthor={Reinhard Heckel},     
    pdfsubject={Subject},   
    pdfcreator={Reinhard Heckel},   
    pdfproducer={Producer}, 
    pdfnewwindow=true,      
    colorlinks=true,       
    linkcolor=red,          
    citecolor=green,        
    filecolor=magenta,      
    urlcolor=cyan           
}

\usepackage{filecontents}

\usepackage[
backend=bibtex,
url=false,isbn=false,doi=false,
date=year,
hyperref=auto,
style=alphabetic,
natbib=true,
sorting=nty,
firstinits=true,
maxnames=2, maxbibnames=10,
]{biblatex}

\bibliography{./bibliography.bib} 

\usepackage{comment}
\usepackage{bm}

\usepackage{tikz}
\usepackage{pgfplots}
\usepgfplotslibrary{groupplots} 
\usetikzlibrary{pgfplots.groupplots}
\usetikzlibrary{matrix,arrows,decorations.pathmorphing}
\usepackage{pgfplots,pgfplotstable}
\usepgfplotslibrary{fillbetween}
\pgfplotsset{compat=1.10}

\usetikzlibrary{matrix,arrows,decorations.pathmorphing}

\usepackage{amsmath,amssymb,ifthen,color,framed,bm}

\usepackage{colortbl}
\definecolor{tblblue}{RGB}{101,124,191}
\definecolor{tblred}{rgb}{1,0.93,0.93}
\definecolor{DarkBlue}{rgb}{0,0,0.7} 
\definecolor{BrickRed}{RGB}{203,65,84}

\newtheorem{lemma}{Lemma}

\newtheorem{theorem}{Theorem}

\newtheorem{proposition}{Proposition}

\newtheorem{assumption}{Assumption}

\newcommand\loss{\mc L}
\newcommand\stepsize{\eta}
\newcommand\iter{t}
\newcommand\iteralt{\tau} 

\newcommand\cn{\mathrm{cn}}

\newcommand\N{N}

\newcommand\mystackrel[2]{\stackrel{\text{#1}}{#2}}

\newcommand\Jac{\mathcal J}

\newcommand\ball{\mc B}
\newcommand\relu{\mathrm{ReLU}}

\newcommand\vtheta{\bm \theta}
\newcommand\mSigma{\bm \Sigma}

\begin{document}

\begin{center}

{\bf{\LARGE{
Compressive sensing with un-trained neural networks: Gradient descent finds the smoothest approximation
}}}

\vspace*{.2in}

{\large{
\begin{tabular}{cccc}
Reinhard Heckel$^\ast$ and Mahdi Soltanolkotabi$^\dagger$
\end{tabular}
}}

\vspace*{.05in}

\begin{tabular}{c}
$^\ast$Dept. of Electrical and Computer Engineering, Technical University of Munich \\
$^\dagger$Dept. of Electrical and Computer Engineering, University of Southern California
\end{tabular}

\vspace*{.1in}

\today

\vspace*{.1in}

\end{center}


\begin{abstract}
Un-trained convolutional neural networks have emerged as highly successful tools for image recovery and restoration. They are capable of solving standard inverse problems such as denoising and compressive sensing with excellent results by simply fitting a neural network model to measurements from a single image or signal without the need for any additional training data. For some applications, this critically requires additional regularization in the form of early stopping the optimization. For signal recovery from a few measurements, however, un-trained convolutional networks have an intriguing self-regularizing property: Even though the network can perfectly fit any image, the network recovers a natural image from few measurements when trained with gradient descent until convergence. In this paper, we provide numerical evidence for this property and study it theoretically. We show that---without any further regularization---an un-trained convolutional neural network can approximately reconstruct signals and images that are sufficiently structured, from a near minimal number of random measurements.
\end{abstract}


\section{Introduction}
Un-trained convolutional neural networks have emerged as highly successful tools for image recovery and restoration, for a variety of problems including denoising, compressive sensing, and inpainting~\cite{ulyanov_deep_2018,Jin_Gupta_Yerly_Stuber_Unser_2019,veen_compressed_2018,jagatap_algorithmic_2019,heckel_regularizing_2019,heckel_deep_2019,Bostan_Heckel_Chen_Kellman_Waller_2020,wang_PhaseImagingUntrained_2020,hyder_GenerativeModelsLowDimensional_2020,arora_untrained_2019}. 
As opposed to trained convolutional neural networks, that learn an image prior from training data, un-trained convolutional networks act as an image prior without any training and solely based on the architecture of the network and the optimization procedure used to fit them.

The benefit of untrained networks was first observed in the Deep Image Prior (DIP) paper~\citep{ulyanov_deep_2018}. The key observation of 
\citet{ulyanov_deep_2018} is that fitting a standard over-parameterized convolutional autoencoder (specifically, the U-net~\citep{ronneberger_u-net_2015} or variations thereoff) to a single noisy/corrupted image, when combined with early stopping, yields excellent denoising, inpainting, and super-resolution performance. 
Subsequent literature has demonstrated that many elements of the architecture of a convolutional autoencoder---such as the encoder part---are irrelevant for this behavior to emerge. In particular the papers~\citep{heckel_deep_2019, heckel_denoising_2020} highlight the critical role of convolutions with fixed convolutional kernels.

\begin{figure*}
\begin{center}
\begin{tikzpicture}
\begin{groupplot}[
y tick label style={/pgf/number format/.cd,fixed,precision=3},
scaled y ticks = false,
legend style={at={(1,1)}},
title style={at={(0.5,3.3cm)}, anchor=south},
group style={group size= 2 by 1, 
xlabels at=edge bottom, 
           horizontal sep=1.3cm, vertical sep=1.0cm}, xlabel={iteration $\iter$},
         width=5.5cm,height=5cm]
	\nextgroupplot[title = {(a) loss},xmode=log,ymode=log,ylabel={MSE}] 
	\addplot +[mark=none] table[x index=1,y index=2]{./fig/CS_convergence_art.dat};
	\addplot +[mark=none] table[x index=1,y index=2]{./fig/CS_convergence_grass.dat};
	\nextgroupplot[title = {(b) loss w.r.t. image},xmode=log,ymode=log] 
	\addplot +[mark=none] table[x index=1,y index=3]{./fig/CS_convergence_art.dat};
	\addplot +[mark=none] table[x index=1,y index=3]{./fig/CS_convergence_grass.dat};
\end{groupplot} 

\begin{scope}[yshift = 1cm]
\node[] at (-5.9cm,3.4cm){
original
};
\node[] at (-3.3cm,3.4cm){
recovered
}; 
\node[draw,red,very thick,inner sep=0.005cm] at (-5.9cm,2cm){
\includegraphics[width=2.2cm]{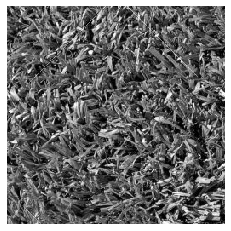}
};
\node[draw,red,very thick,inner sep=0.005cm] at (-3.3cm,2cm){
\includegraphics[width=2.2cm]{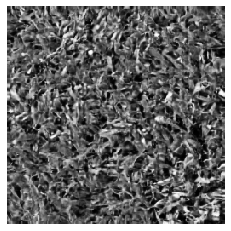}
};
\node[draw,blue,very thick,inner sep=0.005cm] at (-5.9cm,-0.5cm){
\includegraphics[width=2.2cm]{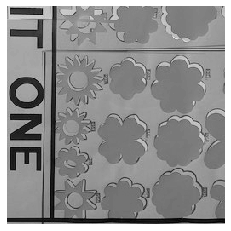}
};
\node[draw,blue,very thick,inner sep=0.005cm] at (-3.3cm,-0.5cm){
\includegraphics[width=2.2cm]{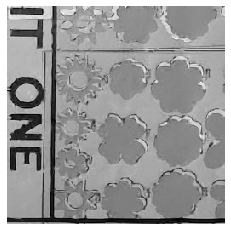}
};
\end{scope}

\end{tikzpicture}
\end{center}
\caption{
Compressive sensing of two different images $\vx^\ast$ displayed on the right with a random matrix $\mA \in \reals^{m\times n}, m = n/4$, from the measurement $\vy = \mA \vx^\ast$.
Panel (a) shows the loss at iteration $\iter$, i.e., $\frac{1}{2}\norm[2]{\mA G(\mC_\iter) - \vy}^2$, and panel (b) is the loss with respect to the original image, i.e., $\norm[2]{G(\mC_\iter) - \vx^\ast}^2$. 
Here, $G$ is a 5-layer deep decoder
\cite{heckel_deep_2019}; a convolutional network with fixed convolutional filters. 
The figure looks qualitatively the same if we take $G$ as the deep image prior~\cite{ulyanov_deep_2018}, a U-net like convolutional autoencoder.
It can be seen that early stopping is not required: gradient descent converges to a good solution, and early stopping does not improve performance for this example. 
Moreover, the simple and smooth image (blue) achieves a smaller loss with the same number of measurements than the non-smooth grass texture (red). Both features are captured by our theory.
\label{fig:csloss}
}
\end{figure*}

Un-trained convolutional networks are empirically most effective when the network is over-parametrized, meaning that is has more parameters than image pixels. This holds even though in this regime the neural network can in principle fit any image perfectly, including random noise. Therefore, further regularization is critical to performance in many applications. For instance  denoising~\cite{ulyanov_deep_2018,heckel_denoising_2020} critically requires early stopping, as without early stopping the noisy image is fitted perfectly and no noise is removed. However, perhaps surprisingly, for some inverse problems including inpainting~\cite{ulyanov_deep_2018} and compressive sensing, no further regularization is necessary! That is, a convolutional neural network, when fitted to compressive measurements from a single image (no other training data) can estimate the original image well, as illustrated in Figure~\ref{fig:csloss}. This phenomenon demonstrates an intriguing self-regularization capability in the context of compressive sensing.

The overarching goal of this paper is to study compressive sensing with un-trained convolutional generators theoretically in order to explain the above phenomenon. In particular, our goal is to understand
(i) why for compressive sensing problems gradient descent can reconstruct a good signal estimate without any further regularization or additional training data and to 
(ii) prove that this is possible with a minimal number of measurements that is proportional to an appropriately defined notion of signal dimensionality.


\subsection{Compressive sensing with un-trained neural networks}

We consider the problem of recovering an unknown signal $\vx^\ast\in\reals^n$ from $m \ll n$ linear measurements of the form
\begin{align}
\label{linm}
\vy = \mA \vx^\ast\in\reals^m,
\end{align}
with $\mA \in \reals^{m \times n}$ representing the measurement matrix. This problem formulation includes the compressive sensing problem relevant for computational imaging as well as inpainting. To understand how un-trained networks can be utilized to recover the unknown signal, consider an over-parameterized, un-trained convolutional image prior  $G\colon \reals^{N} \to \reals^n$ mapping an $N \gg n$ dimensional parameter vector $\mC$ to an $n$ dimensional signal. We take $G$ to be the deep decoder, a simple \emph{un-trained} convolutional network, defined formally in Section~\ref{sec:convgenerator}. 
We emphasize that $G$ is an un-trained neural networks that is randomly initialized and has never seen any training data. 
To reconstruct the signal from its measurements we fit a compressed version of the generator output to these measurements via randomly initialized gradient descent on the loss
\begin{align}
\label{eq:lossintro}
\loss(\mC) = \frac{1}{2}\norm[2]{\mA G(\mC) - \vy}^2.
\end{align}
Let $\hat \mC$ denote the solution found by gradient descent. The signal estimate can then be calculated as $\hat \vx = G(\hat \mC)$. 

A number of recent papers have shown that with the deep image prior (a convolutional autoencoder) or the deep decoder (a convolutional generator) as a prior $G$, this approach is rather effective~\cite{veen_compressed_2018,jagatap_algorithmic_2019,heckel_regularizing_2019}. Most recently Arora et al.~\cite{arora_untrained_2019} have shown that this approach significantly improves upon classical compressive sensing methods ($\ell_1$-regularization and total-variation norm minimization) for accelerating multi-coil magnetic resonance imaging, which is arguably one of the most prominent real-world application of compressive sensing.

The generator $G$ is over-parameterized and can express any image $\vx^\ast$, including unstructured noise.
Nevertheless, typically no further regularization in the form of early stopping the optimization is necessary. We demonstrate this phenomenon in Figure~\ref{fig:csloss}. This figure shows that running gradient descent on the loss $\loss(\mC)$ eventually yields an estimate that is very close to the original image.
This is surprising because i) there is no additional training data and ii) even though the generator $G$ can fit any image, including noise, gradient descent still finds an image close to the original one.

\subsection{Contributions}

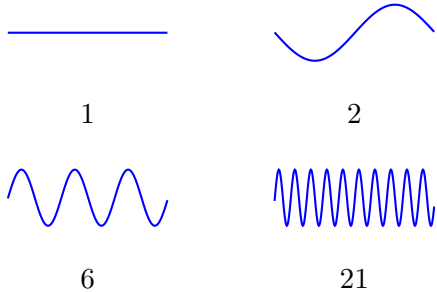
\begin{figure}
\begin{center}
\begin{tikzpicture}
\begin{groupplot}[axis lines=none,
y tick label style={/pgf/number format/.cd,fixed,precision=3},
scaled y ticks = false,
legend style={at={(1.2,1)}},
         title style={at={(0.5,-1.1cm)}, anchor=south}, group
         style={group size= 2 by 2, xlabels at=edge bottom,
           horizontal sep=1cm, vertical sep=1.3cm},
         width=0.25*\textwidth,height=0.15*6.5in]

\nextgroupplot[title={1}]
\addplot +[mark=none,thick] table[x index=0,y index=1]{./fig/circulant_singular_vectors.dat};
\nextgroupplot[title={2}]
\addplot +[mark=none,thick] table[x index=0,y index=2]{./fig/circulant_singular_vectors.dat};
\nextgroupplot[title={6}]
\addplot +[mark=none,thick] table[x index=0,y index=3]{./fig/circulant_singular_vectors.dat};
\nextgroupplot[title={21}]
\addplot +[mark=none,thick] table[x index=0,y index=4]{./fig/circulant_singular_vectors.dat};
\end{groupplot}          
\end{tikzpicture}
\end{center}
\vspace{-0.5cm}
\caption{
\label{fig:circulantsingvectors}
The 1st, 2nd, 6th, and 21st trigonometric basis functions in dimension $n=300$. 
}
\end{figure}

The main contribution of this paper is to show that un-trained convolutional image priors provably enable recovery of natural images from a few random linear measurements. This holds by simply running gradient descent until convergence---without any further regularization.
More specifically, we show that fitting an over-parameterized convolutional network with fixed convolutions (via gradient descent) to random measurements of a smooth signal essentially recovers that signal. Furthermore, the required number of measurements is commensurate to how smooth the signal is with more measurements required when the signal has ``high-frequency" components. 
In more detail:
\begin{itemize}
\item Suppose we have $m$-linear measurements $\vy = \mA \vx^\ast, \mA \in \reals^{m\times n}$ of an unknown signal $\vx^\ast$ with $\mA$ a Gaussian measurement matrix. Furthermore, assume that the signal $\vx^\ast$ is $p$-smooth, in the sense that it can be represented as a linear combination of the $p$ lowest frequency orthonormal trigonometric basis functions $\vw_1,\ldots, \vw_n \in \reals^n$ as
\[
\vx^\ast = \sum_{i=1}^p \vw_i \innerprod{\vw_i}{\vx^\ast}.
\]
We plot these trigonometric basis functions in Figure~\ref{fig:circulantsingvectors} and formally define them later on in Section \ref{main}. Note that the smaller $p$, the smoother the signal $\vx^\ast$ is, thus $p$ is a measure of smoothness.

Our main result shows that the estimate $\mC_\infty$, obtained by running gradient descent on the loss ~\eqref{eq:lossintro} until convergence, yields an output $G(\mC_\infty)$ which is very close to $\vx^\ast$, i.e., $G(\mC_\infty) \approx \vx^\ast$. This holds as soon as the number of measurements exceeds the degrees of smoothness present in the signal ($p$). Since natural images are approximately smooth, this results provides a theoretical explanation why compressive sensing on natural images with over-parameterized convolutional generators works so well (see~\cite{veen_compressed_2018,jagatap_algorithmic_2019,heckel_regularizing_2019,arora_untrained_2019} for corresponding empirical results).

\item In a nutshell, our main insight is that the behavior of large over-parameterized neural networks is dictated by the spectral properties of their Jacobian mapping. For the convolutional generators considered in this paper, the associated Jacobian matrix has singular vectors that can be well approximated by the orthonormal trigonometric basis function and singular values that decay very quickly from the low-frequency to the high-frequency trigonometric basis functions.
Specifically, the associated singular values decay approximately geometrically.

To prove our result, we first characterize the least-squares solution of a randomly sketched least-squares problem with a design matrix with a decaying spectrum. To prove the result for convolutional generators we show that this non-linear learning problem behaves like an associated linear model with the above spectral characteristics. We then conclude the proof for the corresponding convolutional generator, by showing that the solutions obtained by running gradient descent on the non-linear problem is close to that obtained by running gradient descent on the linear problem. 

\item In order to develop a better understanding of compressive sensing with untrained priors, we also carry out compressive sensing experiments for accelerating magnetic resonance imaging (MRI). Our experiments corroborate our theoretical finding that simply iterating until convergence is effective. This also suggests that there is little or no benefit to additional regularization.

\end{itemize}

Our paper is organized as follows: 
We start by stating the convolutional architecture considered in this paper in Section~\ref{sec:convgenerator}. 
In Section~\ref{sec:linear} we study the reconstruction of a signal from few a measurements with a \emph{linear} over-parameterized generator to form intuition. 
In Section~\ref{main} we state our main results for signal recovery with convolutional generators.
Section~\ref{sec:realMRI} contains our numerical result for MRI imaging. 
We conclude the paper with related work and a brief proof sketch, all formal proofs are deferred to the Appendix.

\section{Convolutional generators}
\label{sec:convgenerator}

A convolutional generator generates an image through convolutional operations and applications of non-linearities. In this paper, we study a two-layer convolutional generator $G\colon \reals^{nk \times n} \to \reals^n$ theoretically. The generator has the form
\begin{align}
\label{eq:twodDD}
G(\mC) = \relu(\mU \mC) \vv.
\end{align}
Here, $\vv = [1,\ldots,1, -1,\ldots,-1] / \sqrt{k}$ are the fixed weights of the output layer, of which half are positive and the other half are negative, and $\mC \in \reals^{n \times k}$ is the coefficient matrix of the generator, corresponding to the weights in the first layer of the network.  
Critical for the performance of the generator is the convolutional operation with a fixed kernel $\vu$, implemented through multiplication with the circulant matrix $\mU \in \reals^{n\times n}$.

This architecture is a two-dimensional version of the deep decoder~\cite{heckel_deep_2019}.
The deep decoder in turn is a sub-set of the deep image prior~\cite{ulyanov_deep_2018} and the U-net~\cite{ronneberger_u-net_2015}, as commented on below.

The deep decoder with $d$ layers (typically, $d=4,5,6$) is defined as
\begin{align}
\label{eq:deepdecoder}
G(\mC)
=
\relu( \mU \mB_d \mC_d )\vv,
\end{align}
where 
\begin{align*}
\mB_{i+1} 
= 
\cn(\relu(\mU_i\mB_i \mC_i )), i = 0,\ldots, d-1.
\end{align*}
Here $\cn(\cdot)$ is a channel normalization operation, which normalizes each channel/column of the volume/matrix $\relu(\mU_i\mB_i \mC_i) \in \reals^{n_i\times k}$ individually and can be viewed as a special case of the batch normalization operation. 
Note that if the signal to be generated is an image and thus two-dimensional ($n_i \in \mathbb Z^2$), then $\mB_i$ is a three-dimensional tensor consisting of $k$ many channels, and if the signal is one-dimensional ($n_i \in \mathbb Z$), those tensors are two-dimensional and can be viewed as matrices consisting of $k$ many columns (or channels). 
Moreover, $\mB_0$ is a fixed input tensor, which we assume to have full row rank. 
The parameters of the deep decoder are the weight matrices $\mC_1,\ldots,\mC_d \in \reals^{k\times k}$.
Multiplication with those weight matrices is performing linear combinations of the channels, which in turn is equivalent to performing 1x1-convolutions. 

For $d=2$, the deep decoder reduces to the two-dimensional version in~\eqref{eq:twodDD}.
To see this, note that for $d=2$, because $\mB_0$ has full column rank, optimizing over $\mB_0\mC_0 \in \reals^{n \times k}$ is equivalent to optimizing over $\mC \in \reals^{n \times k}$ instead. 

Finally, as mentioned before, the deep decoder can be viewed as the relevant part of a convolutional generator to function as an image prior. It can be deduced from a convolutional autoencoder (such as the deep image prior~\cite{ulyanov_deep_2018} and the U-net~\cite{ronneberger_u-net_2015}) by removing the encoder part, any skip connections, and most surprisingly, the trainable convolutional filters of spatial extent larger than one. As demonstrated in~\cite{heckel_denoising_2020}, the critical aspect for an un-trained deep image prior are the convolutions with fixed convolutional kernels, implemented here by the operator $\mU$. 


\section{Signal recovery with over-parameterized linear generators 
}
\label{sec:linear}

Consider an over-parameterized linear generator $\tilde G(\vc) = \mJ \vc$ defined by a wide, full-rank, generator matrix $\mJ \in \reals^{n \times N}, N \geq n$, and an arbitrary and unknown signal $\vx^\ast \in \reals^n$. 
Because $\mJ$ has full rank, the signal can be expressed as $\vx^\ast = \mJ \vc^\ast$. 
However, the coefficient vector $\vc^\ast$ in this representation is non-unique, as $\mJ$ is a wide matrix containing more columns than rows. 
We observe $m$ linear measurements of the unknown signal of the form
\[
\vy = \mA \vx^\ast,
\]
where $\mA \in \reals^{m\times n}$ is a wide ($m < n$) Gaussian measurement matrix, with iid $\mc N(0,1/m)$ entries. We note that with this variance, norms are approximately preserved (i.e., for a fixed $\vz$, with high probability $\norm[2]{\vz} \approx \norm[2]{\mA \vz}$). 

Our goal is to estimate the signal $\vx^\ast$ based on the measurement $\vy$. We estimate the signal $\vx^\ast$ by first computing a coefficient estimate $\hat \vc$ by minimizing the loss
\[
\loss(\vc) = \frac{1}{2} \norm[2]{\mA \mJ \vc - \vy}^2,
\]
via running gradient descent with sufficiently small step size until convergence. We then estimate the signal via 
$\hat \vx = \mJ \hat \vc$. 
Since gradient descent applied on a least-squares problem yields the minimum-norm solution, the estimate $\hat \vc$ can equivalently be expressed as
\begin{align}
\label{eq:defchat}
\hat \vc = \arg \min_\vc \norm[2]{\vc}^2
\text{ subject to } \mA \mJ \vc = \vy.
\end{align}
In closed form, $\hat \vc$ is given as
\[
\hat \vc 
= \pinv{(\mA \mJ)} \mA \mJ \vc^\ast
= \mP_{ \mJ^T \mA^T} \vc^\ast,
\]
where $\pinv{(\mA \mJ)}$ is the pseudo-inverse of $\mA \mJ$, and $\mP_{ \mJ^T \mA^T }$ is a orthogonal projection operator onto the range of $\transp{(\mA \mJ)}$. 
Thus, the signal estimation error is
\begin{align}
\label{eq:esterr}
\hat \vx - \vx^\ast
=
\mJ (\hat \vc - \vc^\ast) 
= 
\mJ ( \mI - \mP_{ \mJ^T \mA^T } ) \vc^\ast.
\end{align}
The following theorem characterizes this signal estimation error.

\begin{theorem}
\label{prop:specialcase}
Let $\mA \in \reals^{m \times n}$ be a random Gaussian matrix with $m \geq 12$, and let $\vw_1,\ldots,\vw_n$ be the left singular vectors of $\mJ$ with associated singular values $\sigma_1\geq \ldots \geq \sigma_n$.
Then, for any $\vx^\ast \in \reals^n$, with probability at least $1- 3e^{-1/2 m}$, the signal estimate $\hat \vx = \mJ \hat \vc$ based on the measurement $\vy = \mA \vx^\ast$, with the coefficient estimate $\hat \vc(\vy)$ defined in~\eqref{eq:defchat}, obeys
\begin{align}
\label{eq:boundthmlinear}
\norm[2]{\hat \vx - \vx^\ast}^2
\leq
C
\left( \sum_{i=1}^n \frac{1}{\sigma_i^2} \innerprod{\vw_i}{\vx^\ast}^2 \right)
 \sum_{i > 2m/3 } \sigma_i^2.
\end{align}
Here, $C$ is a fixed numerical constant.
\end{theorem}

The proof, given in the appendix, relies on arguments from~\citep[Sec.~8 and Sec.~9]{halko_finding_2011} developed for approximating low-rank matrices through random sampling.

The theorem guarantees that the error in estimating the signal $\vx^\ast$ from compressive measurements $\vy = \mA \vx^\ast$ is small provided that two conditions are satisfied:
\begin{enumerate}
\item[(i)] The signal $\vx^\ast$ lies (approximately) in the span of the leading $O(m)$ singular vectors of $\mJ$, where $m$ is the number of linear measurements.
\item[(ii)] The singular values of the generator matrix $\mJ$ decay sufficiently fast (for example geometrically).
\end{enumerate}
To see this, let us consider a concrete example.
Suppose the singular values decay geometrically, i.e., 
$\sigma_i^2 = \gamma^i$ for some $\gamma \in (0,1)$. Moreover, suppose that the signal $\vx^\ast$ lies in the span of the leading $m/3$ singular values of $\mJ$, i.e., $\vx^\ast \in \mathrm{span}(\vw_1,\ldots, \vw_{m/3})$. 
Then, Theorem~\ref{prop:specialcase} guarantees that the estimate $\hat \vx$ based on $m$ random linear measurements obeys
\begin{align}
\label{eq:simplifiedbound}
\norm[2]{\hat \vx - \vx^\ast}^2
\leq
C\frac{\gamma^{m/3}}{1-\gamma}
\norm[2]{\vx^\ast}^2.
\end{align}
Here, we used that the first term in the right-hand-side of~\eqref{prop:specialcase} is bounded by $1/\sigma_{m/3}^2 \norm[2]{\vx^\ast}^2$, using that $\vx^\ast$ is in the span of the leading singular vectors, and that $\sum_{i>2m/3} \sigma_i^2 \leq \frac{\gamma^{2m/3}}{1-\gamma}$, by the formula for a geometric series. 
The bound~\eqref{eq:simplifiedbound} is very small provided that $\gamma$ is slightly below one (since $\gamma^{m/3}$ decays exponentially)---thus guaranteeing almost perfect recovery of a signal that is aligned with the leading singular vectors of $\mJ$.

\section{Main results for compressive sensing with convolutional generators}
\label{main}
We are now ready to state our main results for compressive sensing with convolutional generators. We consider the non-linear least-squares objective
\[
\loss(\mC) = \frac{1}{2} \norm[2]{\mA G(\mC) - \vy}^2,
\]
where $\mA \in \reals^{m\times n}, m \leq n$, is a Gaussian random matrix with iid $\mc N(0,1/m)$ entries and $G(\mC)$ is the two-layer decoder network  defined in section~\ref{sec:convgenerator}. 
We minimize this objective by running gradient descent with a constant stepsize $\eta$, starting from a random initialization $\mC_0$, with entries drawn iid from a Gaussian distribution $\mc N(0,\omega^2)$, and with variance $\omega^2$ specified later. The coefficients at iterations $\iter = 1,2, \ldots$ are given by
\begin{align}
\mC_{\iter+1} = \mC_\iter - \eta \nabla \loss(\mC_\iter).
\end{align}
In the previous section we studied a linear generator with generator matrix $\mJ$ with quickly decaying spectrum.
In this section we extend the insights from the previous section to the non-linear case by replacing the role of the generator matrix $\mJ$ with the Jacobian of the non-linear generator $G$, defined as $[\Jac(\mC)]_{ij} = \frac{\partial}{\partial c_i} [G(\mC)]_j$. In contrast to the linear case, however, the Jacobian changes across iterations of gradient descent. Nevertheless, we can account for these changes in the Jacobian in our analysis.

\begin{figure}
\begin{center}
\begin{tikzpicture}
\begin{groupplot}[
y tick label style={/pgf/number format/.cd,fixed,precision=3},
scaled y ticks = false,
legend style={at={(1,1)}},
         title style={at={(0.5,-1.3cm)}, anchor=south}, group
         style={group size= 4 by 1, xlabels at=edge bottom, 
           horizontal sep=1.2cm, vertical sep=1.5cm},
         width=0.27*6.5in,height=0.27*6.5in]
\nextgroupplot[title = {(a) triangular kernel},xmin=120,xmax=180] 
\addplot +[mark=none,thick] table[x index=0,y index=2]{./fig/triangles_kernels.dat};
\addplot +[mark=none,thick] table[x index=0,y index=3]{./fig/triangles_kernels.dat};
\nextgroupplot[title = {(b) associated weights},ymode=log,xmode=log]
\addplot +[mark=none,thick] table[x index=0,y index=2]{./fig/triangles_spectra.dat};
\addplot +[mark=none,thick] table[x index=0,y index=3]{./fig/triangles_spectra.dat};
\end{groupplot}          
\end{tikzpicture}
\end{center}
\vspace{-0.5cm}
\caption{
\label{fig:kernels}
Triangular kernels and the weights associated to low-frequency trigonometric functions they induce, for a generator network of output dimension $n=300$. The wider the kernel is, the more the weights are concentrated towards the low-frequency components of the signal. Note that the lower singular values decay geometrically (as evident from the straight line in the log-log plot)---as the singular values in our example in Section~\ref{sec:linear}.  
}
\end{figure}
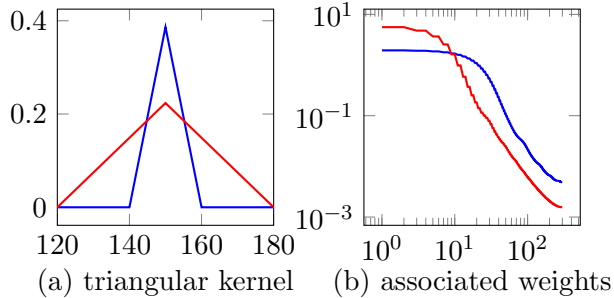

As found in~\cite{heckel_denoising_2020}, for the two-layer deep decoder that we consider, the left singular vectors of the Jacobian can be well approximated by the trigonometric basis function $\vw_1,\ldots,\vw_n\in \reals^n$ plotted in Figure~\ref{fig:circulantsingvectors}, and defined as
\begin{align}
\label{eq:trigonfunc}
[\vw_{i}]_j
=
\frac{1}{\sqrt{n}}
\begin{cases}
1 & i = 0 \\
\sqrt{2} \cos( 2\pi ji/n) & i = 1,\ldots,n/2-1 \\
(-1)^{j} & i =n/2 \\
\sqrt{2} \sin(2\pi ji/n) & i = n/2+1,\ldots,n-1
\end{cases}.
\end{align}
Moreover, the singular values of the Jacobian throughout the iterates can be well approximated by associated values that only depend on the convolution kernel $\vu$ associated with the convolution operator $\mU$. Those values  $\bm \sigma \in \reals^n$ are given by
\begin{align}
\label{eq:singvals}
\bm \sigma
=
\norm[2]{\vu}
\sqrt{\Bigg|\mtx{F}g\left(\frac{\vu\circledast\vu}{\norm[2]{\vu}^2}\right)\Bigg|}
\end{align}
with
\begin{align*}
g(z)=\frac{1}{2}\left(1-\frac{\cos^{-1}\left(z\right)}{\pi}\right)z.
\end{align*}
 Here, for two vectors $\vu,\vv \in \reals^n$, $\vu\circledast\vv$ denotes their circular convolution, $\mF$ is the discrete Fourier transform matrix, and the scalar non-linearity $g$ is applied entrywise.
As a concrete relevant example, in Figure~\ref{fig:kernels} we depict the triangular kernel that is used in the original deep decoder network. 
The most important observation from this plot is that the associated weights $\bm \sigma = [\sigma_1, \ldots, \sigma_n]$ decay very fast, namely geometrically.

With those definition, we are now ready to state our main result. 

\begin{theorem}
\label{thm:noearlystopping}
Let $\mA \in \reals^{m\times n}$ be a random Gaussian matrix with $m \geq 12$ and suppose we are given a linear measurement $\vy = \mA \vx^\ast$ of an arbitrary signal $\vx^\ast \in \reals^n$.
Consider a two layer generator network $G(\mC) = \relu(\mU \mC) \vv$, $\mC \in \reals^{n\times k}$, with
\begin{align}
\label{kreq}
k
\ge 
C_{\vu} \frac{m}{\xi^8},
\end{align}
channels and with convolutional kernel $\vu$ of the convolutional operator $\mU$ and associated weights $\bm \sigma = [\sigma_1, \ldots, \sigma_n]$. Here, $\xi\le 1$ is arbitrary and $C_{\vu}$ is a constant that only depends on the convolutional kernel $\vu$. In order to estimate the signal, we fit the convolutional generator to the signal by running gradient descent starting from a random initialization $\mC_0$ with i.i.d.~$\mathcal{N}(0,\omega^2)$, entries, $\omega \propto \frac{\norm[2]{\vy}}{\sqrt{n} }$, 
and sufficiently small stepsize 
to the loss $\frac{1}{2}\norm[2]{\mA G(\mC) - \vy}^2$ until convergence. 
Then, with high probability, the reconstruction error with parameters $\mC_\infty$ at convergence obeys 
\begin{align}
\label{eq:reccond}
\norm[2]{G(\mC_\infty) - \vx^\ast}^2
\leq&
C
\left( \sum_{i=1}^n \frac{1}{\sigma_i^2} \innerprod{\vw_i}{\vx^\ast}^2 \right)
\sum_{i > 2m/3} \sigma_i^2 
 + \xi^2\norm[2]{\vx^\ast}^2.
\end{align}
Here, $C$ is a fixed numerical constant.
\end{theorem}

Theorem~\ref{thm:noearlystopping} establishes that a convolutional generator enables the reconstruction of a natural signal from a few linear measurements. 
To see this, note that a good model for a natural image is a smooth signal, i.e., a signal that can be well-approximated by few leading trigonometric basis functions. 
More concretely, Figure 4 in \citep{Simoncelli_Olshausen_2001} shows that the power spectrum of a natural image (i.e., the energy distribution by frequency) decays rapidly from low frequencies to high frequencies. 

Thus it is reasonably to assume that the signal $\vx^\ast$ can be represented with few of the trigonometric basis function; for concreteness say that $\vx^\ast$ lies in the span of $\vw_1,\ldots,\vw_{m/3}$. 
Next, recall from Figure~\ref{fig:kernels} that the weights associated with a triangular kernel decay geometrically (i.e., $\sigma_i^2 = \gamma^i$ for some $\gamma \in (0,1)$). Thus, from the same argument as used for~\eqref{eq:simplifiedbound}, the bound~\eqref{eq:reccond} established by the theorem yields that the reconstruction error is bounded by
\begin{align*}
\norm[2]{G(\mC_\infty) - \vx^\ast}^2
\leq
C\frac{\gamma^{m/3}}{1-\gamma}
\norm[2]{\vx^\ast}^2 + \xi^2\norm[2]{\vx^\ast}^2.
\end{align*}
Thus our theorem guarantees the recovery of a sufficiently smooth signal by optimizing over the range of the generator. 
In particular if the signal is $p$-smooth, i.e., lies in the span of $\vw_1,\ldots, \vw_p$, then $O(p)$ measurements are sufficient to provide an accurate estimate.


\subsection{Beyond two layer networks}

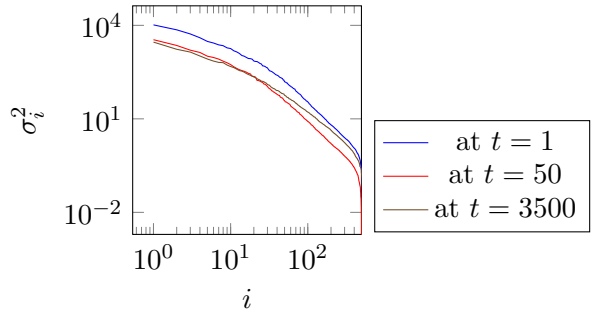
\begin{figure}
\begin{center}
\begin{tikzpicture}
\begin{groupplot}[
y tick label style={/pgf/number format/.cd,fixed,precision=3},
scaled y ticks = false,
legend columns=1, 
        legend style={
                    at={(2.0,0.5)},
        },
         title style={at={(1.0cm,-0.55cm)}, anchor=south}, group
         style={group size= 1 by 1, xlabels at=edge bottom, ylabels at=edge left,yticklabels at=edge left,
           horizontal sep=0.4cm, vertical sep=1.4cm}, xlabel={$i$}, ylabel={$\sigma_i^2$},
         width=0.28*6.5in,height=0.28*6.5in,
         ]

\nextgroupplot[ymode=log, xmode=log, title style={at={(1.0cm,-2.2cm)}, anchor=south}, xmax=496,] 
	\addplot +[mark=none] table[x index=0,y index=1]{./fig/JacobianSingularVal.dat};
	\addlegendentry{at $t=1$}
	\addplot +[mark=none] table[x index=0,y index=2]{./fig/JacobianSingularVal.dat};
	\addlegendentry{at $t=50$}
	\addplot +[mark=none] table[x index=0,y index=3]{./fig/JacobianSingularVal.dat};
	\addlegendentry{at $t=3500$}
	
\end{groupplot}    
\end{tikzpicture}
\end{center}
\vspace{-0.6cm}
\caption{
\label{fig:jacobiandeep}
The singular value distribution of the Jacobian of a four-layer deep decoder at different iterations of gradient descent; the spectrum changes only slightly, and the singular values decay slightly faster than geometrically.
}
\end{figure}

Our main theorem from the previous section relies on two critical ingredients:
\begin{enumerate}
\item[(i)] The finding from~\cite{heckel_denoising_2020} that the leading singular vectors of the Jacobian of a two-layer deep decoder are approximately the trigonometric basis function throughout all iterations of gradient descent.
\item[(ii)] The weights $\sigma_1,\ldots,\sigma_n$ associated with the trigonometric basis functions decaying sufficiently fast, specifically approximately geometric. 
That is required for gradient descent applied to fitting $m$ compressive measurements until convergence to (approximately) only fit the signal to the leading $O(m)$ trigonometric basis functions.
\end{enumerate}
Those results extend to deeper networks as follows. 
First, as shown numerically in~\cite{heckel_denoising_2020}, the leading singular vectors of the Jacobian of a four-layer deep decoder are also close to the trigonometric basis functions, and change only little across iterations. 
Second, as shown in Figure~\ref{fig:jacobiandeep}, the singular values of a four-layer deep decoder also decay (at least) geometrically, and the spectrum changes only little across iterations.  
Thus, the implications of our theory continue to apply for deeper deep decoders.

\section{Numerical experiments for magnetic resonance imaging}
\label{sec:realMRI}

In the final part of our paper we consider accelerating magnetic resonance imaging (MRI), one of the major application of compressive sensing. 
MRI is a medical imaging technique where measurements of an object can only be taken in the Fourier domain, referred to as $k$-space.
If the full $k$-space measurement is collected, an image of the object can be computed almost perfectly (up the noise inherent in the measurement process). In order to accelerate the imaging process, it is common to only collect a small part of the $k$-space, which corresponds to taking few linear Fourier measurements; or in the notation of our paper, a measurement matrix $\mA$ with subsampled rows of the Fourier matrix.

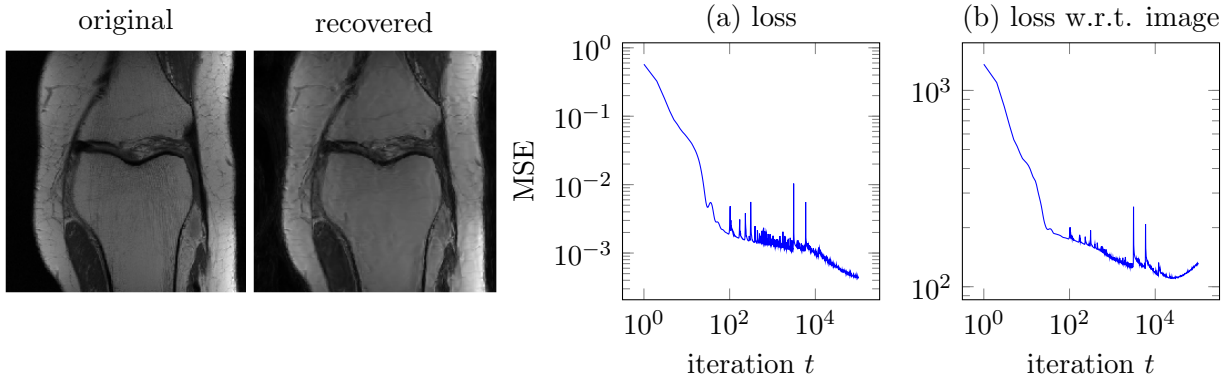
\begin{figure*}
\begin{center}
\begin{tikzpicture}
\begin{groupplot}[
y tick label style={/pgf/number format/.cd,fixed,precision=3},
scaled y ticks = false,
legend style={at={(1,1)}},
title style={at={(0.5,3.2cm)}, anchor=south},
group style={group size= 3 by 1, 
xlabels at=edge bottom, 
           horizontal sep=1.1cm, vertical sep=2.4cm}, xlabel={iteration $\iter$},
         width=5cm,height=5cm]
	\nextgroupplot[title = {(a) loss},xmode=log,ymode=log,ylabel={MSE}] 
	\addplot +[mark=none] table[x index=1,y index=2]{./fig/CS_convergence_MRI.dat};
	\nextgroupplot[title = {(b) loss w.r.t. image},xmode=log,ymode=log] 
	\addplot +[mark=none] table[x index=1,y index=3]{./fig/CS_convergence_MRI.dat};
\end{groupplot}

\node at (-6.6cm,3.7cm){
original
};
\node at (-3.3cm,3.7cm){
recovered
};

\node at (-6.6cm,1.7cm){
\includegraphics[width=3.4cm]{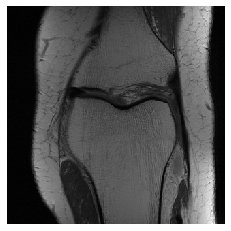}
};
\node at (-3.3cm,1.7cm){
\includegraphics[width=3.4cm]{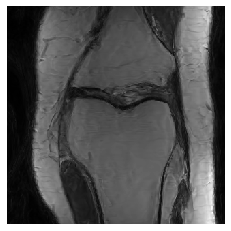}
};

\end{tikzpicture}
\end{center}
\caption{
Compressive sensing MRI:
MSE of reconstructing an image from 8-fold undersampled k-space MRI measurements. While early stopping is not absolutely necessary, stopping at about 2000 iterations slightly improves performance relative to optimizing until convergence.
\label{fig:csMRI}
}
\end{figure*}

In order to understand whether our main finding---that signal reconstruction from compressive measurements without further regularization is possible---applies in practice, we consider the problem of reconstructing an image from few k-space measurements.
We consider reconstruction of an image from 8-fold undersampled k-space measurements from the fastMRI dataset, recently released by facebook and NYU~\cite{zbontar_fastMRI_2018}. 
We reconstruct with a $d=5$ layer and highly over-parameterized deep decoder. 
Figure~\ref{fig:csMRI} shows the corresponding loss curves. 
It can be seen that early stopping at the optimal early stopping point gives only marginally better performance than when optimizing until convergence, and in addition the optimal early stopping point is unknown in practice (because we do not have access to a reconstruction from a full measurement). 

\section{Related literature}
In this paper we focus on un-trained neural network for solving inverse problems. In contrast a large body of recent result concentrates on using trained deep convolutional neural networks for image recovery and reconstruction. 
Training based deep learning methods for solving inverse problems are either trained end-to-end for tasks like denoising \citep{burger_image_2012,zhang_beyond_2017}, or are based on learning a generative image model (by training an  autoencoder or GAN~\citep{hinton_reducing_2006,goodfellow_generative_2014}) and then using the resulting image models to regularize problems such as compressed sensing \citep{bora_compressed_2017, hand_global_2017,huang_ProvablyConvergentScheme_2018}, denoising~\citep{heckel_deep_2018}, or phase retrieval~\citep{handleongvoroninski2019, shamshad2018robust}. 
In contrast to un-trained network, where optimization is over the weights of the un-trained generator, in the aformentioned papers it is over the input of the (trained) network.

Our proof relies on relating the dynamics of gradient descent on an over-parameterized network to that of gradient descent on an associated linear network. 
This proof technique has been used in a variety of recent publication~\citep{soltanolkotabi2018theoretical, venturi2018spurious, du_gradient_2018,Oymak:2018aa,oymak_towards_2019,arora_fine-grained_2019, oymak_generalization_2019,basri_convergence_2019,li_gradient_2019}. 
Most related to our work is the recent paper~\cite{heckel_denoising_2020} that shows that the deep decoder enables denoising. 
Neither of the publications, however, addresses compressive sensing or reconstruction from randomly sketched data, and most of our technical results are specific to this setup.

Finally note that regularizing \emph{linear} models with gradient descent via early stopping has a rich history in the signal processing community. In the 50s, Landweber proposed to recover a signal from linear measurements via gradient descent~\cite{Landweber_1951} which became known as the Landweber algorithm in the inverse problems community. Subsequent work in this literature proposed to early-stop the Landweber iterations (i.e., gradient descent) in order to regularize ill-posed inverse problems~\cite{Trussell_Civanlar_1985}.

\section{Proof sketch}

In this section we provide a sketch of our argument. 
Our statement and formal proof pertains to the two-layer case, in this section we provide the sketch for the general case where $G(\vtheta)$ is a generic network with a $N$-dimensional parameter vector $\vtheta$, and then comment on how this general proof strategy is particularized to the two layer case.

Given a measurement $\vy$, we characterize the solution of running gradient descent with fixed step size $\eta$ on the nonlinear least-squares objective
\[
\loss(\vtheta) = \frac{1}{2} \norm[2]{ f(\vtheta) - \vy}^2,
\quad f(\vtheta) = \mA G(\vtheta), 
\]
starting from an initial point $\vtheta_0$. The updates take the form
\begin{align}
\label{iterupdates}
\vtheta_{\iter + 1} = \vtheta_\iter - \stepsize \nabla \loss(\vtheta_\iter),
\quad
\nabla \loss(\vtheta)
= \transp{\Jac}(\vtheta) ( f(\vtheta) - \vy),
\end{align}
where $\Jac(\vtheta)$ is the Jacobian of $f$ at $\vtheta$. 
We start gradient descent from a random initialization $\vtheta_0$ with iid $\mc N(0,\omega)$ entries. 
Central to our analysis are the following objects. 
Let $\Jac_G(\vtheta) \in \reals^{n\times N}$ be the Jacobian of $G(\vtheta)$ and define $\mJ_G$ as a reference generator Jacobian that we set to a matrix that is very close to the generator Jacobian at initialization, i.e., $\mJ_G \approx \Jac_G(\vtheta_0)$. 
For the two-layer network for which we state a precise result, this matrix only depends on the convolutional operator $\mU$.

Relevant for the dynamics of gradient descent, however, are the corresponding sketched original and reference Jacobians, defined as
\[
\Jac(\vtheta) = \mA \Jac_G(\vtheta) \in \reals^{m \times N}
\quad
\text{and}
\quad
\mJ = \mA \mJ_G \in \reals^{m \times N}.
\]
Since we chose $\mJ_G \approx \Jac_G(\vtheta_0)$, we also have $\mJ \approx \Jac (\vtheta_0)$.

\subsection{Closeness to an associated linear problem}
\newcommand\losslin{\loss_{\mathrm{lin}}}
\newcommand\flin{f_{\mathrm{lin}}}

To characterized the behavior of the gradient descent updates in~\eqref{iterupdates}, we relate the non-linear least squares problem to a linearized one in a ball around the initialization $\vtheta_0$. 
This general strategy has been utilized in a number of recent publications~\citep{soltanolkotabi2018theoretical, du_gradient_2018,arora_fine-grained_2019,oymak_towards_2019,oymak_generalization_2019,heckel_denoising_2020}. 
We define the associated linearized least-squares problem as
\begin{align}
\label{linprob}
\losslin(\vtheta)
=
\frac{1}{2}
\norm[2]{ f(\vtheta_0) + \mJ (\vtheta - \vtheta_0) - \vy}^2.
\end{align}
Starting from the same initial point $\vtheta_0$, the gradient descent updates of the linearized problem are
\begin{align}
\label{iterupdateslin}
\widetilde{\vtheta}_{\iter + 1}
&= 
\widetilde{\vtheta}_\iter - \stepsize \transp{\mJ} \left( f(\vtheta_0) + \mJ (\widetilde{\vtheta}_\iter - \vtheta_0) - \vy\right). 
\end{align}

The iterates and residuals of the non-linear and linear updates are close throughout the entire run of gradient descent provided the following assumptions are satisfied:
\begin{enumerate}
\item[(i)] The smallest and largest singular values of the
generator reference Jacobian are lower and upper bounded by constants $\alpha$ and $\beta$, respectively. 
\item[(ii)] The reference Jacobian approximates the Jacobian at initialization, i.e., for $\epsilon_0>0$, 
\begin{align*}
\norm{\mJ - \Jac(\vtheta_0)} \leq \epsilon_0,
\end{align*}
where $\norm{\cdot}$ is the standard operator (matrix) norm.
\item[(iii)] Within a radius $R$ around the initialization, the Jacobian varies by no more than $\epsilon$ in the sense that
\begin{align}
\label{eq:Jacclose}
\norm{\Jac(\vtheta) - \Jac(\vtheta_0)}
\leq
\frac{\epsilon}{2},
\quad 
\text{for all}
\quad
\vtheta \in \ball_R(\vtheta_0).
\end{align}
Here, $\ball_R(\vtheta_0) \defeq \{\vtheta \colon \norm[2]{\vtheta - \vtheta_0} \leq R\}$ is the ball with radius $R$ around $\vtheta_0$.
\end{enumerate}
Under these assumptions, we establish that the residuals of the linear problem,
\[
\widetilde{\vr}_\iter \defeq 
f(\vtheta_0) + \mA \mJ_G (\tilde \vtheta_\iter - \vtheta_0) - \vy
\] 
and that of the non-linear problem,
\[
\vr_\iter \defeq \mA G(\vtheta_\iter) - \vy,
\]
are close during the entire run of gradient descent, and most importantly for proving our result, that the iterates of the linear and non-linear problem are close, again during the entire run of gradient descent:
\begin{align*}
\norm[2]{\vtheta_\iter - \widetilde{\vtheta}_\iter} 
&\leq 
O(\epsilon_0 + \epsilon)\norm[2]{\vr_0}.
\end{align*}

\subsection{Inheriting the properties of the linear problem}

Recall that our goal is to characterize the signal estimate $G(\vtheta_\infty)$ at convergence.
We characterize this estimate by
\begin{enumerate}
\item[i)] characterizing the estimate $\hat \vx = \mJ_G \vtheta_\infty$ obtained by running the linear problem until convergence and
\item[ii)] showing that this estimate is close to the original estimate, i.e., $\mJ_G \vtheta_\infty \approx G(\vtheta_\infty)$. 
\end{enumerate}

In more detail, suppose that the assumption i-iii are satisfied for sufficiently small closeness parameters $\epsilon_0$ and $\epsilon$. 
Then, as discussed above, the iterates of the non-linear problem and the linear problem are close at any iteration, in particular at convergence. Since the Jacobians are also close, we can establish that $\hat \vx = \mJ_G \vtheta_\infty \approx G(\vtheta_\infty)$. 

In more detail, we can bound the signal estimation error at convergence as
\begin{align*}
\norm[2]{G(\vtheta_\infty) - \vx^\ast}
&\leq
\norm[2]{\hat \vx  - \vx^\ast}
+
\norm[2]{G(\vtheta_\infty)   - \hat \vx} \\
&\leq
\norm[2]{\hat \vx  - \vx^\ast} 
+ 
O(\epsilon_0 + \epsilon).
\end{align*}
The first term is controlled by analyzing the linear case with Theorem~\ref{prop:specialcase} from Section~\ref{sec:linear}. To control the second term we need a simple definition
\begin{align*}
\mathcal{J}_G(\mtx{\vtheta}_\infty,0)=\int_0^1 \mathcal{J}_G(t\mtx{\vtheta}_\infty) dt.
\end{align*}
With this definition in place we can proceed to bound the second term  as follows
{\small
\begin{align*}
&\norm[2]{G(\vtheta_\infty)   - \hat \vx} \\
&=
\norm[2]{ \Jac_G(\vtheta_\infty,0) \vtheta_\infty  - \mJ_G \tilde \vtheta_\infty} \\
&=
\norm[2]{ \Jac_G(\vtheta_\infty,0) \vtheta_\infty - \Jac_G(\vtheta_\infty,0) \tilde \vtheta_\infty + \Jac_G(\vtheta_\infty,0) \tilde \vtheta_\infty  - \mJ_G \tilde \vtheta_\infty} \\
&\leq
\norm{ \Jac_G(\vtheta_\infty,0)} \norm[2]{ \vtheta_\infty - \tilde \vtheta_\infty}
+ 
\norm{\Jac_G(\vtheta_\infty,0) - \mJ_G} \norm[2]{\tilde \vtheta_\infty} \\
&\leq O(\epsilon_0 + \epsilon).
\end{align*}
}
For the last bound we used that by our discussion above, the iterates of the non-linear problem are close at any iteration, in particular at convergence, so that $\norm[2]{ \vtheta_\infty - \tilde \vtheta_\infty} \leq O(\epsilon_0 + \epsilon)$.

\subsection{Concluding the proof sketch}

The proof for the two-layer case is then concluded by
analyzing the associated linear problem. 
In particular, we use that the matrix $\mJ_G$ has as its left-singular vectors the trigonometric basis function, and its spectrum are the associated weights $\sigma_1,\ldots,\sigma_n$ specified in Section~\ref{main}.

In order to extend this proof to a multi-layer deep decoder $G(\vtheta)$, all we need to do is to characterize the associated matrix $\mJ_G$, in particular its left-singular vectors and corresponding singular values.

\section*{Code}
Code to reproduce the experiments is available at 
\url{https://github.com/MLI-lab/cs_deep_decoder}.
\section*{Acknowledgements}
R. Heckel is partially supported by NSF award IIS-1816986 and acknowledges support of the NVIDIA Corporation in form of a GPU. 
M. Soltanolkotabi is supported by the Packard Fellowship in Science and Engineering, a Sloan Research
Fellowship in Mathematics, an NSF-CAREER under award \#1846369, the Air Force Office of Scientific
Research Young Investigator Program (AFOSR-YIP) under award \#FA9550-18-1-0078, an NSF-CIF award
\#1813877, DARPA under the Learning with Less Labels (LwLL) and Fast Network Interface Cards (FastNICs) program, and a Google faculty research award.

\printbibliography

\newpage

\appendix

\appendix

\section{Proof of Theorem~\ref{prop:specialcase}}

The statement follows from the following more general result.

\begin{proposition}
\label{prop:generalcase}
Let $\mA \in \reals^{m \times n}$ be a Gaussian random matrix with $m = k+p$, and $p\geq 4$, and let $\transp{\mJ} = \mU_n \mSigma \transp{\mV}$
with $\mU_n \in \reals^{d\times n}$ and $\mSigma, \mV \in \reals^{n \times n}$,
be the singular value decomposition of $\transp{\mJ}$ with singular values $\sigma_1\geq \ldots \geq \sigma_n$. 
Then, for any $\vc^\ast \in \reals^d$, with probability at least $1- 2e^{-p} - e^{-u^2/2}$, the estimate 
$\hat \vc = \mP_{\transp{\mA} \transp{\mJ}} \vc$ obeys
\[
\norm[2]{\mJ \hat \vc - \mJ \vc}^2
\leq
\norm[2]{\transp{\mU}_n \vc^\ast}^2
\left(
\left(
\sigma_{k+1} e \left( 
\sqrt{\frac{3k}{p+1}}
+ 
\frac{ e \sqrt{k+p} }{p+1} u
\right)
+
\sqrt{\sum_{i>k} \sigma_i^2 }
\frac{e \sqrt{k+p} }{p+1} u
\right)^2
+
\sum_{j > k} \sigma_j^2
\right).
\]
\end{proposition}

To see this, note that with $p=k/2$ and $u=\sqrt{p}$, the proposition guarantees that with probability at least $1 - 3e^{-p}$,
\begin{align*}
\norm[2]{\mJ \hat \vc - \mJ \vc^\ast}^2
&\leq
\norm[2]{\transp{\mU}_n \vc^\ast}^2
\left(
\left(
25 \sigma_{k+1} 
+ 7\sqrt{\sum_{i>k} \sigma_i^2 }
\right)^2
+
\sum_{i > k} \sigma_i^2 \right) \\
&\leq
\norm[2]{\transp{\mU}_n \vc^\ast}^2
32^2 \sum_{i>k} \sigma_i^2.
\end{align*}
Noting that $m = 3/2 k$, $\hat \vx = \mJ \hat \vc$ and $\vx^\ast = \mJ \vc^\ast$ concludes the proof.

\paragraph{Proof of Proposition~\ref{prop:generalcase}:}
By the characterization~\eqref{eq:esterr}, our goal is to upper bound 
\begin{align}
\norm[2]{\mJ \hat \vc - \mJ \vc^\ast}^2
=
\norm[2]{\transp{\vc^\ast} ( \mI - \mP_{ \mJ^T \mA^T } ) \transp{\mJ} }^2.
\end{align}
Our proof relies on arguments from~\citep[Sec.~8 and Sec.~9]{halko_finding_2011} developed for approximating low-rank matrices through random sampling. 

We start by partitioning the right-singular vectors of $\transp{\mJ}$ into two blocks $\mV_1$ and $\mV_2$ containing $k$ and $n-k$ columns, respectively. 
\[
\transp{\mJ} = \mU_n 
\begin{bmatrix}
\mSigma_1 & 0 \\
0 & \mSigma_2 
\end{bmatrix}
\begin{bmatrix}
\transp{\mV}_1 \\
\transp{\mV}_2
\end{bmatrix}.
\]
Define the random matrices
\[
\Omega_1 = \transp{\mV}_1 \transp{\mA} \in \reals^{k \times m},
\quad 
\Omega_2 = \transp{\mV}_2 \transp{\mA} \in \reals^{n-k \times m}.
\]
Note that both matrices are standard Gaussian, and, because they are non-overlapping sub-matrices of $\mV \mA$, they are also stochastically independent. 
Moreover, $\Omega_1$ has full row-rank with probability one. 

For convenience, define
\[
\transp{\tilde \mJ} = \mSigma \mV.
\]
Next, we record a useful property from~\citep[Prop.~8.4]{halko_finding_2011}: For a unitary matrix $ \mU$ any matrix $\mM$, 
\begin{align}
\mP_{\mM} = \mU \mP_{ \transp{\mU} \mM} \transp{\mU}.
\label{eq:projemtrx}
\end{align}
To see that the identity~\eqref{eq:projemtrx} holds, first note that the matrix $\mP = \transp{\mU} \mP_{\mM} \mU$ is an orthogonal projection operator because it is Hermitian an $\mP^2 = \mP$. Moreover,
\[
\mathrm{range}(\mP)  
= \transp{\mU} \mathrm{range}(\mM)
= \mathrm{range}(\transp{\mU}\mM).
\]
Since the range determines the orthogonal projector onto its range, we have that $\mP = \transp{\mU} \mP_{\mM} \mU = \mP_{\transp{\mU}\mM}$, concluding the proof of~\eqref{eq:projemtrx}. 
Next, let 
\[
\transp{\mJ} 
= 
\underbrace{[\mU_n \mU_{d-n}]}_{\mU}
\begin{bmatrix}
\mSigma_1 & 0 \\
0 & \mSigma 2 \\
0  & 0
\end{bmatrix}
\transp{\mV}
\]
be the full singular value decomposition of $\transp{\mJ}$, including the singular vectors $\mU_{d-n}$ multiplying with zero singular values.
Applying the identity~\eqref{eq:projemtrx} and that 
$\transp{\mU} \mU$
we proceed as
\begin{align*}
\norm[2]{\transp{\vc} ( \mI - \mP_{ \mJ^T \mA^T } ) \transp{\mJ} }^2
&=
\norm[2]{\transp{\vc} \mU ( \mI - \mP_{\transp{\mU} \mJ^T \mA^T } ) \transp{\mU} \transp{\mJ} }^2 \\
&=
\norm[2]{\transp{\vc} [\mU_n \mU_{d-n}] ( \mI - \mP_{ \begin{bmatrix} \tilde \mJ^T \mA^T \\ 0 \end{bmatrix} } ) \begin{bmatrix} \mSigma \\ 0 \end{bmatrix}  }^2 \\
&=
\norm[2]{\transp{\vc} [\mU_n \mU_{d-n}] \begin{bmatrix}(\mI - \mP_{ \tilde \mJ^T \mA^T })\mSigma \\ 0 \end{bmatrix}  ) }^2 \\
&=
\norm[2]{\transp{\vc} \mU_n ( \mI - \mP_{ \tilde \mJ^T \mA^T } )  \mSigma }^2.
\end{align*}
Moreover,
\begin{align*}
\norm[2]{\transp{\vc} \mU_n ( \mI - \mP_{ \tilde \mJ^T \mA^T } )  \mSigma }^2
&\leq
\norm[2]{\transp{\vc} \mU_n}^2 
\norm{ ( \mI - \mP_{ \tilde \mJ^T \mA^T } )  \mSigma }^2 \\
&=
\norm[2]{\transp{\vc} \mU_n}^2 
\norm{\transp{\mSigma} ( \mI - \mP_{ \tilde \mJ^T \mA^T } )  \mSigma} \\
&\leq
\norm{ \mSigma_2 \Omega_2 \pinv{\Omega}_1 }^2 + \norm{\mSigma_2}^2 \\
&\leq
\norm[2]{\transp{\vc} \mU_n}^2 
\left(
\left(
\norm{\mSigma_2} e \left( 
\sqrt{\frac{3k}{p+1}}
+ 
\frac{ e \sqrt{k+p} }{p+1} u
\right)
+
\norm[F]{\mSigma_2}
\frac{e \sqrt{k+p} }{p+1} u t
\right)^2
+ \norm{\mSigma_2}^2
\right),
\end{align*}
where the second-to-last inequality follows from~\citep[Last ineq in Sec.~9.2]{halko_finding_2011}.
Finally, the last inequality holds with the probability specified in the proposition because by~\citep[Last inequality in Sec.~10.3]{halko_finding_2011}, for $p \geq 4$ and $u>0$,
\begin{align*}
\PR{ \norm{ \mSigma_2 \Omega_2 \pinv{\Omega}_1 } 
\geq
\norm{\mSigma_2} e \left( 
\sqrt{\frac{3k}{p+1}}
+ 
\frac{ e \sqrt{k+p} }{p+1} u
\right)
+
\norm[F]{\mSigma_2}
\frac{e \sqrt{k+p} }{p+1} u t
}
\leq 2 e^{-p} + e^{-u^2/2} .
\end{align*}
This concludes the proof of the proposition.



\section{Proof of Theorem~\ref{thm:noearlystopping}}

The result stated in the main text (Theorem~\ref{thm:noearlystopping}) is obtained from a slightly more general result which applies beyond convolutional networks. 
Specifically, we consider neural network generators of the form
\begin{align*}
G(\mC) = \relu( \mU \mC) \vv,
\end{align*}
with $\mC\in\reals^{n\times k}$, and $\mU\in\reals^{n\times n}$ an arbitrary fixed matrix, and $\vv\in\reals^k$, with half of the entries of $\vv$ equal to $+1/\sqrt{k}$ and the other half equal to $-1/\sqrt{k}$.

The (transposed) Jacobian $\relu(\mU \vc)$ is $\transp{\mU} \diag( \relu'(\mU \vc) ) $.
Thus the Jacobian of $G(\mC)$ is given by
\begin{align}
\label{eq:JacmC}
\transp{\Jac}_G(\mC)
=
\begin{bmatrix}
v_1 \transp{\mU} \diag( \relu'(\mU \vc_1) ) \\
\vdots \\
v_k \transp{\mU} \diag( \relu'(\mU \vc_k)) \\
\end{bmatrix}
\in \reals^{nk \times n},
\end{align}
where $\relu'$ is the derivative of the activation function. 
Next we define a notion of expected Jacobian. 
Towards this goal, we first define the matrix 
\begin{align*}
\mSigma(\mU)
\defeq
 \EX{\Jac_G(\mC)\Jac^T_G(\mC)},
\end{align*}
associated with the function $G(\mC) = \relu( \mU \mC) \vv$. 
Here, expectation is over $\mC$ with iid $\mc N(0,\omega)$ entries.
Consider the eigenvalue decomposition of $\mSigma(\mU)$ given by
\[
\mSigma(\mU) =\sum_{i=1}^n \sigma_i^2\vw_i\vw_i^T.
\]
Our results depend on the largest and smallest eigenvalue of $\mSigma(\mU)$ denoted by $\sigma_n^2$ and $\norm{\mU}^2$ and in particular a condition number denoted by $\kappa$ formally defined as
\begin{align*}
\kappa_{\vu}:=\frac{\norm{\mtx{U}}^2}{\sigma_n^2}.
\end{align*}
With these definitions in place we are now ready to state our result about neural generators.
\begin{theorem}
\label{thm:mainResult}
Consider a compressive observation $\vy\in\reals^m$ given by
\[
\vy = \mA \vx,
\]
where $\mA \in \reals^{m\times n}$ with $m\le \frac{n}{9}$ is a Gaussian random matrix with iid $\mc N(0,1/m)$ entries. 
Suppose that the number of channels obeys
\begin{align}
\label{eq:thmoverparamass}
k
\ge 
 C\frac{\kappa_{\vu}^{26}}{\xi^8}m
\end{align}
for an error tolerance parameter $0<\xi \le \frac{1}{\sqrt{2\log\left(\frac{2n}{\delta} \right)}}$. 
We fit the neural generator $G(\mC)$ to the signal $\vy \in\reals^n$ by minimizing a loss of the form 
\begin{align}
\label{lossc}
\mathcal{L}(\mC)=\frac{1}{2}\norm[2]{\mA G(\mC)-\vy}^2
\end{align}
via running gradient descent with iterations $\mC_{\iter+1} = \mC_\iter- \stepsize\nabla\mathcal{L}(\mC_\iter)$, starting from $\mC_0$ with i.i.d.~$\mathcal{N}(0,\omega^2)$ entries, $\omega =  \frac{\xi\norm[2]{\vy} }{2\sqrt{n}\norm{\mU}}$, and step size obeying $\eta\le \frac{m}{4n\norm{\mtx{U}}^2}$. Then, with probability at least $1-ne^{-k^2}-2e^{-\frac{m}{2}}-\delta$, for all iterations $\iter$,
\begin{align}
\label{finalconcc}
\norm[2]{\vx - G(\mC_\iter)}
&\leq \xi\norm[2]{\vx^\ast}
+
C \left( \sum_{i=1}^n \frac{1}{\sigma_i^2} \innerprod{\vw_i}{\vx^\ast}^2 \right)
\sum_{i > 2m/3} \sigma_i^2
.
\end{align}
\end{theorem}

Theorem~\ref{thm:noearlystopping} follows directly from Theorem~\ref{thm:mainResult} by noting that for $\mU$ a circulant matrix (implementing a convolution), as found in~\cite{heckel_denoising_2020}, the left singular vectors of $\mSigma(\mU)$ are given by the trigonometric basis functions in~\eqref{eq:trigonfunc} and the singular values are given by~\eqref{eq:singvals}.


\section{The dynamics of linear and nonlinear least-squares}\label{metathms}

Theorem~\ref{thm:mainResult}, proven below, builds on a result on the dynamics of a general non-linear least squares problem that is stated and discussed in this section. 
Consider a nonlinear least-squares fitting problem of the form
\[
\loss(\vtheta) = \frac{1}{2} \norm[2]{ f(\vtheta) - \vy}^2.
\]
Here, $f\colon \reals^\N \rightarrow \reals^n$ is a non-linear model with parameters $\vtheta \in \reals^\N$. 

To solve this problem, we run gradient descent with a fixed stepsize $\stepsize$, starting from an initial point $\vtheta_0$, with updates of the form
\begin{align}
\label{iterupdates}
\vtheta_{\iter + 1} = \vtheta_\iter - \stepsize \nabla \loss(\vtheta_\iter)
\quad \text{where}\quad
\nabla \loss(\vtheta)
= \transp{\Jac}(\vtheta) ( f(\vtheta) - \vy).
\end{align}
Here, $\Jac(\vtheta)\in\reals^{n\times N}$ is the Jacobian associated with the nonlinear map $f$ with entries given by
$
[\Jac(\vtheta)]_{i,j} = \frac{\partial f_i(\vtheta)}{ \partial \vtheta_j }
$. 
In order to study the properties of the gradient descent iterates in~\eqref{iterupdates}, we relate the non-linear least squares problem to a linearized one in a ball around the initialization $\vtheta_0$. 
This general strategy has been utilized in a variety of recent publications~\citep{soltanolkotabi2018theoretical, du_gradient_2018,arora_fine-grained_2019,oymak_towards_2019,oymak_generalization_2019}, our specific argument is most similar to~\cite{heckel_denoising_2020}. 
Contrary to the result in~\cite{heckel_denoising_2020}, which holds for a certain number of initial iterations, our statement applied to all iterations. 

The associated linearized least-squares problem is defined as
\begin{align}
\label{linprob}
\losslin(\vtheta)
=
\frac{1}{2}
\norm[2]{ f(\vtheta_0) + \mJ (\vtheta - \vtheta_0) - \vy}^2.
\end{align}
Here, $\mJ\in\reals^{n\times \N}$, refered to as the reference Jacobian, is a fixed matrix independent of the parameter $\vtheta$ that approximates the Jacobian mapping at initialization, $\Jac(\vtheta_0)$. 
Starting from the same initial point $\vtheta_0$, the gradient descent updates of the linearized problem are
\begin{align}
\label{iterupdateslin}
\widetilde{\vtheta}_{\iter + 1}
&= 
\widetilde{\vtheta}_\iter - \stepsize \transp{\mJ} \left( f(\vtheta_0) + \mJ (\widetilde{\vtheta}_\iter - \vtheta_0) - \vy\right).
\end{align}
To show that the non-linear updates~\eqref{iterupdates} are close to the  linearized iterates~\eqref{iterupdateslin}, we make the following assumptions:
\begin{assumption}[Bounded spectrum]\label{BndSpec} We assume the singular values of the reference Jacobian obey for some $\alpha,\beta$
\begin{align}
\label{eq:Jacbounded0}
\sqrt{2} \alpha \le \sigma_{n} \le \sigma_{1} \le \beta.
\end{align}
Furthermore, we assume that the Jacobian mapping associated with the nonlinear model $f$ obeys
\begin{align}
\label{eq:Jacbounded}
\norm{\Jac(\vtheta)} \leq \beta\quad\text{for all}\quad \vtheta\in\reals^N.
\end{align}
\end{assumption}
\begin{assumption}[Closeness of the reference and initialization Jacobians]\label{assJac2} We assume the reference Jacobian and the Jacobian of the nonlinearity at initialization $\Jac(\vtheta_0)$ are $\epsilon_0$-close in the sense that
\begin{align}
\label{eq:assJacJclose}
\norm{\Jac(\vtheta_0) - \mJ} \leq \epsilon_0.
\end{align}
\end{assumption}

\begin{assumption}[Bounded variation of Jacobian around initialization]\label{assJac3}
We assume that within a radius $R$ around the initialization, the Jacobian varies by no more than $\epsilon$ in the sense that
\begin{align}
\label{eq:Jacclose}
\norm{\Jac(\vtheta) - \Jac(\vtheta_0)}
\leq
\frac{\epsilon}{2},
\quad 
\text{for all}
\quad
\vtheta \in \ball_R(\vtheta_0),
\end{align}
where $\ball_R(\vtheta_0) \defeq \{\vtheta \colon \norm{\vtheta - \vtheta_0} \leq R\}$ is the ball with radius $R$ around $\vtheta_0$.
\end{assumption}
Under these assumptions
i) the difference of the nonlinear iterative updates \eqref{iterupdates} and the linear iterative updates \eqref{iterupdateslin} is bounded, and 
ii) the difference of the linear and non-linear residuals, defined as 
\begin{align}
&\text{nonlinear residual:}\quad \vr_\iter \defeq f(\vtheta_\iter) - \vy\label{twores1}\\
&\text{linear residual:}\quad\quad\text{ }\text{ }
\widetilde{\vr}_\iter \defeq
f(\vtheta_0) + \mJ (\widetilde{\vtheta}_\iter - \vtheta_0) - \vy
\label{twores2}
\end{align}
are close throughout the entire run of gradient descent; both  in the proximity of the initialization.

\begin{theorem} [Closeness of linear and nonlinear least-squares problems]
\label{thm:maintechnical}
Assume the Jacobian mapping $\Jac(\vtheta)\in\reals^{n\times \N}$ associated with the function $f(\vtheta)$ obeys Assumptions \ref{BndSpec}, \ref{assJac2}, and \ref{assJac3} around an initial point $\vtheta_0\in\reals^\N$ with respect to a reference Jacobian $\mJ\in\reals^{n\times \N}$ and with parameters $\alpha, \beta, \epsilon_0, \epsilon$, obeying 
$2\beta (\epsilon_0 + \epsilon) \leq \alpha^2$, and $R$. Furthermore, assume the radius $R$ is given by
\begin{align}
\label{eq:defR}
\frac{R}{2} 
\defeq
\norm[2]{ \mJ^\dagger\vr_0}
+2.5\frac{\beta^2}{\alpha^4}(\epsilon_0 + \epsilon)   \norm[2]{\vr_0}.
\end{align}
Here, $\pinv{\mJ}$ is the pseudo-inverse of $\mJ$. 
We run gradient descent with stepsize $\stepsize \leq \frac{1}{\beta^2}$ on the linear and non-linear least squares problem, starting from the same initialization $\vtheta_0$. 
Then, for all iterations $\iter$,
\begin{enumerate}
\item[i)] the non-linear residual converges geometrically
\begin{align}
\label{resnl}
\norm[2]{\vr_{\iter}}\le \left(1-\eta\alpha^2\right)^t \norm[2]{\vr_0},
\end{align}
\item[ii)]  the residuals of the original and the linearized problems are close
\begin{align}
\norm[2]{\vr_\iter - \widetilde{\vr}_\iter} 
&\leq
2 \beta \eta (\epsilon_0 + \epsilon)(1 - \eta \alpha^2)^{\iter-1}
 \iter \norm[2]{\vr_0} \label{eq:propvrkclose}\\
&\leq
\frac{2\beta (\epsilon_0 + \epsilon)}{e(\ln{2}) \alpha^2}
\norm[2]{\vr_0},\label{eq:propvrkclose2}
\end{align}
\item[iii)]  the parameters of the original and the linearized problems are close
\begin{align}
\norm[2]{\vtheta_\iter - \widetilde{\vtheta}_\iter} 
&\leq 
2.5 \frac{\beta^2}{\alpha^4} (\epsilon_0 + \epsilon)\norm[2]{\vr_0},
\label{eq:propthetakclose}
\end{align}
\item[iv)]  and finally, 
the parameters are not far from the initialization 
\begin{align}
\label{eq:indhypinRBall}
\norm[2]{\vtheta_\iter - \vtheta_0} \leq \frac{R}{2}.
\end{align}
\end{enumerate}
\end{theorem}
The above theorem formalizes that in a (small) radius around the initialization, the non-linear problem behaves similarly as its linearization.
Thus to characterize the dynamics of the nonlinear problem, it suffices to characterize the dynamics of the linearized problem. This is the subject of our next theorem, which is a standard results on the iterates of least squares, see \citep[Thm.~5]{heckel_denoising_2020} for the proof.
%
\begin{proposition}[{Theorem~5 in \cite{heckel_denoising_2020}}]
\label{lem:rk}
Consider a linear least squares problem~\eqref{linprob} and let $\mJ = \mW \mSigma \transp{\mV} \in \reals^{n\times p}=\sum_{i=1}^n\sigma_i \vw_i \vv_i^T$
be the singular value decomposition of the matrix $\mJ$.  Then the residual $\widetilde{\vr}_\iter$ after $\iter$ iterations of gradient descent with updates~\eqref{iterupdateslin} is
\begin{align}
\label{eq:linresformula}
\widetilde{\vr}_\iter = \sum_{i=1}^n \left(1 - \stepsize \sigma^2_i\right)^\iter \vw_i \innerprod{\vw_i}{\vr_0}.
\end{align}
Moreover, using a step size satisfying $\stepsize \leq \frac{1}{\sigma_1^2}$, the linearized iterates \eqref{iterupdateslin} obey
\begin{align}
\label{eq:lindifftheta}
\norm[2]{\widetilde{\vtheta}_\iter - \vtheta_0}^2
=
\sum_{i=1}^n
\left(
\innerprod{\vw_i}{\vr_0}
\frac{1 - (1 - \stepsize \sigma^2_i)^\iter}{\sigma_i}
\right)^2.
\end{align}
\end{proposition}
In the next section we show we can combine these two general theorems to provide guarantees for compressed sensing using general neural networks.



\subsection{Proof of Theorem \ref{thm:maintechnical} (closeness of linear and non-linear least-squares)}

The proof is by induction. We note that the base case $\iter=0$ is trivially true. We suppose the statement, in particular the bounds~\eqref{resnl}, \eqref{eq:propvrkclose}, \eqref{eq:propvrkclose2}, \eqref{eq:propthetakclose}, and \eqref{eq:indhypinRBall} hold for all iterations $\iteralt \leq \iter-1$. 
We then show that those relations continue to hold for iteration $\iter$ in five steps: 
In Step I, we show that a weaker version of~\eqref{eq:indhypinRBall} holds, specifically that $\norm[2]{\vtheta_\iter-\vtheta_0}\le R$. 
This guarantees that we can work with our assumptions; those require the iterates to be sufficiently close to the initial values. In Step II we show that the nonlinear residual decreases at a geometric rate proving \eqref{resnl}. In Steps III and IV we show that the residuals and the coefficients of the linear and non-linear problem are close, respectively. Finally, in Step V we utilize Steps I-IV to complete the proof by showing that the iterates of the non-linear problem are close to its initialization (i.e., equation~\eqref{eq:indhypinRBall}).

\paragraph{Linear convergence of linear residual:}
Before we start, we note that under our assumption, the residual of the linear problem converges linearly. Specifically, by the updates of the linear problem~\eqref{iterupdateslin}, we have that
\begin{align}
\label{eq:residual_linear}
\widetilde{\vr}_{\iter + 1} = (\mI - \eta \mJ \transp{\mJ}) \widetilde{\vr}_\iter. 
\end{align}
Using that the smallest singular values of $\mJ \transp{\mJ}$ is lower bounded by $2\alpha^2$, this guarantees that
\begin{align*}
\norm[2]{\widetilde{\vr}_{\iter}}
\leq
(1- 2\eta  \alpha^2)^\iter \norm[2]{\widetilde{\vr}_0},
\end{align*}
establishing linear convergence of the linear problem. 

\paragraph{Step I: Next iterate obeys $\vtheta_\iter \in \ball_R(\vtheta_0)$.} 
We start by using a coarse argument that establishes $\vtheta_\iter \in \ball_R(\vtheta_0)$.  First note that by the triangle inequality and the induction assumption \eqref{eq:indhypinRBall} we have
\begin{align*}
\norm[2]{\vtheta_\iter - \vtheta_0}
\leq&
\norm[2]{\vtheta_\iter - \vtheta_{\iter-1}}
+
\norm[2]{\vtheta_{\iter-1} - \vtheta_0},\\
\leq& \norm[2]{\vtheta_\iter - \vtheta_{\iter-1}}+\frac{R}{2}.
\end{align*}
So to prove $\norm[2]{\vtheta_\iter-\vtheta_0}\le R$ it suffices to show that $\norm[2]{\vtheta_{\iter} - \vtheta_{\iter-1}} \le R/2$. To this aim note that
\begin{align}
\label{myeq}
\frac{1}{\stepsize}\norm[2]{\vtheta_{\iter} - \vtheta_{\iter-1}}
&= \norm[2]{\nabla \loss(\vtheta_{\iter-1})} \nonumber \\
&= 
\norm[2]{ \transp{\Jac}(\vtheta_{\iter-1}) \vr_{\iter-1} } \nonumber \\
&\leq
\norm[2]{ \transp{\Jac}(\vtheta_{\iter-1})  \widetilde{\vr}_{\iter-1} } 
+
\norm{ \Jac(\vtheta_{\iter-1})} 
\norm[2]{  \vr_{\iter-1} - \widetilde{\vr}_{\iter-1}}
\nonumber \\
&\leq
\norm[2]{ \transp{\mJ}  \widetilde{\vr}_{\iter-1} } 
+
\norm{\Jac(\vtheta_{\iter-1}) - \mJ}
\norm[2]{  \widetilde{\vr}_{\iter-1} }
+
\norm{ \Jac(\vtheta_{\iter-1})} 
\norm[2]{\vr_{\iter-1} - \widetilde{\vr}_{\iter-1}} 
\nonumber \\
&\mystackrel{(i)}{\leq} 
\beta^2\norm[2]{ \pinv{\mJ} \vr_{0} } 
+
(\epsilon + \epsilon_0)
\norm[2]{ \vr_0 }
+\frac{2\beta^2 (\epsilon_0 + \epsilon)}{e(\ln{2}) \alpha^2}
\norm[2]{\vr_0}\nonumber\\
&\mystackrel{(ii)}{\leq} 
\beta^2\norm[2]{ \pinv{\mJ} \vr_{0} } 
+\frac{2\beta^2 }{ \alpha^2}(\epsilon_0 + \epsilon)
\norm[2]{\vr_0}.
\end{align}
Here, (ii) follows from the fact that $\frac{1}{2}\le \frac{\beta^2}{\alpha^2}$ and inequality (i) follows from Assumptions \ref{BndSpec}-\ref{assJac3}, the induction hypothesis~\eqref{eq:propvrkclose2}, $\norm{\widetilde{\vr}_{\tau-1}} \leq \norm{\vr_0}$, and the bound
\begin{align*}
\norm[2]{ \transp{\mJ} \widetilde{\vr}_{\iter-1} }
&=
\norm[2]{\transp{\mJ} (\mI - \stepsize \mJ \transp{\mJ})^{\iter-1} \vr_0 } \\
&= \norm[2]{\mSigma (\mI - \stepsize \mSigma^2)^{\iter-1} \transp{\mW}\vr_0} \\
&\le \sqrt{\sum_{j=1}^n\sigma_j^2\langle \vw_j,\vr_0\rangle^2} \\
&\le \beta^2\sqrt{\sum_{j=1}^n\frac{1}{\sigma_j^2}\langle \vw_j,\vr_0\rangle^2} \\
&= \beta^2\norm[2]{\pinv{\mJ}\vr_0}.
\end{align*} 
To continue we use the fact that $\eta\le \frac{1}{\beta^2}$ in \eqref{myeq} to conclude that
\begin{align}
\norm[2]{\vtheta_{\iter} - \vtheta_{\iter-1}}
&\leq
\stepsize\beta^2\norm[2]{ \pinv{\mJ} \vr_{0} } 
+
\stepsize\frac{2 \beta^2 (\epsilon_0 + \epsilon)}{\alpha^2}
\norm[2]{\vr_0}.
\nonumber \\
&\leq
\norm[2]{ \pinv{\mJ} \vr_{0} } 
+\frac{2 (\epsilon_0 + \epsilon)}{\alpha^2}
\norm[2]{\vr_0}.
\nonumber \\
&\leq  \frac{R}{2}. \nonumber
\end{align}
The last inequality follows by definition of $R$ in~\eqref{eq:defR}, and concludes the proof of Step I. 

\paragraph{Step II: 
Geometric decay of non-linear iterate.
} 
Since the linear residuals converge linearly and the Jacobian of the non-linear problem is close the Jacobian of the linear problem, $\mJ$, the non-linear problem also converges linearly. 
To see this, with $\Jac(\va,\vb) = \int_0^1 \Jac(s \vb - (1-s) \va) ds$, we have that, by the mean value theorem
\begin{align*}
f(\vtheta_{\iter}) 
&=
f(\vtheta_{\iter-1} - \eta \nabla \loss(\vtheta_{\iter-1})) \\
&=
f(\vtheta_{\iter-1}) - \eta \Jac(\vtheta_{\iter},\vtheta_{\iter-1}) \nabla \loss(\vtheta_{\iter-1}) \\
&=
f(\vtheta_{\iter-1}) - \eta \Jac(\vtheta_{\iter},\vtheta_{\iter-1}) \transp{\Jac}(\vtheta_{\iter-1}) (f(\vtheta_{\iter-1}) - \vy) \\
&=
f(\vtheta_{\iter-1}) - \eta \mB_1 \mB_2 (f(\vtheta_{\iter-1}) - \vy).
\end{align*}
where in the last equality we defined the matrices $\mB_1$ and $\mB_2$ accordingly for notational convenience. 
This implies that
\begin{align}
\vr_{\iter}
&=
f(\vtheta_{\iter}) - \vy \nonumber \\
&= (\mI - \eta \mB_1 \mB_2) (f(\vtheta_{\iter-1}) - \vy) \nonumber \\
&= (\mI - \eta \mB_1 \mB_2) \vr_{\iter-1}.
\label{eq:iteratesnonlinear}
\end{align}
Thus,
\begin{align*}
\norm[2]{\vr_{\iter}}
&\leq \norm{\mI - \eta \mB_1 \mB_2} \norm[2]{\vr_{\iter-1}} \\
%
%
&\leq \left( \norm{\mI - \eta \mJ \transp{\mJ}}  + \eta \norm{ \mJ \transp{\mJ} - \mB_1 \mB_2} \right)\norm[2]{\vr_{\iter-1}} \\
&\mystackrel{(i)}{\leq} \left( 1 - 2 \eta \alpha^2 + 2\eta \beta (\epsilon_0 + \epsilon) \right)\norm[2]{\vr_{\iter-1}} \\
&\mystackrel{(ii)}{\leq} \left( 1 - \eta \alpha^2 \right)\norm[2]{\vr_{\iter-1}}
\end{align*}
For inequality (ii) we used the assumption $2\beta (\epsilon_0 + \epsilon) \leq \alpha^2$, and for inequality (i) we used the bound
\begin{align}
\norm{\mJ \transp{\mJ} - \mB_1 \mB_2}
&=
\norm{\mJ \transp{\mJ} - \mJ \mB_2 + \mJ \mB_2 - \mB_1 \mB_2} \nonumber \\
&\leq
\norm{\mJ} \norm{ \transp{\mJ} - \mB_2 } 
+
\norm{ \mJ  - \mB_1} \norm{ \mB_2} 
\leq 2 \beta (\epsilon_0 + \epsilon),
\label{eq:boundJJBB}
\end{align}
where the last inequality follows from our assumptions, and using that, by the triangle inequality and assumptions~\ref{assJac2} and \ref{assJac3}, we have
\begin{align}
\label{eq:bounddiffjacbs}
\norm{ \mB_2 - \transp{\mJ}} 
=
\norm{\Jac(\vtheta_{\iter-1} ) - \mJ} 
\leq
\norm{\Jac(\vtheta_{\iter-1} ) - \Jac(\vtheta_0)}
+
\norm{\Jac(\vtheta_0) - \mJ}
\leq
\epsilon_0 + \epsilon.
\end{align}
This establishes that
\begin{align}
\label{eq:residuals_conv}
\norm[2]{ \vr_{\iter}}
\leq
(1- \eta \alpha^2) \norm[2]{\vr_{\iter-1}}\le(1- \eta \alpha^2) ^t\norm[2]{\vr_0} , 
\end{align}
where in the last inequality we used the induction hypothesis \eqref{resnl}. This completes the proof of the bound~\eqref{resnl} for iteration $t$ concluding Step II.
\paragraph{Step III: Original and linearized residuals are close.}
In this step, we bound the deviation of the residuals of the original and linearized problem defined as
\[
\ve_{\iter} \defeq \widetilde{\vr}_{\iter} - \vr_{\iter}.
\]
Specifically, we use the induction hypothesis together with the fact that based on Step I we have $\vtheta_{\iter-1},\vtheta_{\iter } \in \ball_R(\vtheta_0)$, to show
that
\begin{align}
\label{eq:growthperturbations}
\norm{\ve_{\iter}}
\leq
2 \beta \eta (\epsilon_0 + \epsilon) (1 - \eta \alpha^2)^{\iter-1}
 \iter \norm[2]{\vr_0}.
\end{align}
Before we prove this however note that for $x\le 1/2$ we have $(1-x)^{t-1}t\le \frac{1}{e(\ln 2) x}$ for all $t\ge 0$. Now using this identity with $x=\eta\alpha^2\le \frac{\alpha^2}{\beta^2}\le \frac{1}{2}$ in \eqref{eq:growthperturbations} we conclude that
\begin{align*}
\norm{\ve_{\iter}}
&\leq
\frac{2\beta (\epsilon_0 + \epsilon)}{e(\ln 2) \alpha^2}
\norm[2]{\vr_0},
\end{align*}
completing the proof of \eqref{eq:propvrkclose2} for iteration $t$. Thus, all that remains in this step is to establish~\eqref{eq:growthperturbations}. To this aim note that from the formulas for the linear and non-linear residuals in~\eqref{eq:residual_linear} and \eqref{eq:iteratesnonlinear}, we have that
\[
\widetilde{\vr}_{\iter} = (\mI - \eta \mJ \transp{\mJ}) \widetilde{\vr}_{\iter-1}. 
\]
Thus for $\ve_{\iter}= \widetilde{\vr}_{\iter } - \vr_{\iter}$ we have, with the same notation as in step II,
\begin{align*}
\norm{\ve_{\iter}}
&=
\norm[2]{
(\mI - \eta \mJ \transp{\mJ}) \widetilde{\vr}_{\iter-1}
-
(\mI - \eta \mB_1 \mB_2) \vr_{\iter-1}
} \\
&=
\norm[2]{
(\mI - \eta \mJ \transp{\mJ}) (\widetilde{\vr}_{\iter-1} - \vr_{\iter-1})
+
\eta (\mB_1 \mB_2 - \mJ\transp{\mJ}) \vr_{\iter-1}
} \\
&\leq
\norm{\mI - \eta \mJ \transp{\mJ}} \norm[2]{\widetilde{\vr}_{\iter-1} - \vr_{\iter-1}}
+
\eta \norm{\mB_1 \mB_2 - \mJ\transp{\mJ}} \norm[2]{\vr_{\iter-1}} \\
&\leq
(1 - \eta \alpha^2) \norm[2]{\ve_{\iter-1}}
+
2 \eta \beta (\epsilon_0 + \epsilon ) (1 - \eta \alpha^2)^{\iter-1} \norm[2]{\vr_0},
\end{align*}
where the last inequality follows from
$\norm{\mB_1 \mB_2 - \mJ\transp{\mJ}} \leq 2 \beta (\epsilon_0 + \epsilon)$, by~\eqref{eq:boundJJBB}, and 
from using the fact that $\norm[2]{\vr_{\iter-1}} \leq (1 - \eta \alpha^2)^{\iter-1} \norm[2]{\vr_0}$ which holds based on Step II. Finally, plugging in the induction hypothesis $\norm[2]{\ve_{\iter-1}}\le c \xi^{\iter-2}(t-1)\norm[2]{\vr_0}$ with $\xi:=1-\eta\alpha^2$ and $c:=2\eta\beta(\epsilon_0+\epsilon)$ in the above we conclude that 
\begin{align*}
\norm{\ve_{\iter}}\le& \xi\norm{\ve_{\iter-1}}+c\xi^{t-1}\norm[2]{\vr_0}\\
\le& c \xi^{\iter-1}(t-1)\norm[2]{\vr_0}+c\xi^{t-1}\norm[2]{\vr_0}\\
=&c\xi^{t-1}t\norm[2]{\vr_0}\\
=&2\eta\beta(\epsilon_0+\epsilon)\left(1-\eta\alpha^2\right)^{t-1}t\norm[2]{\vr_0}.
\end{align*}
%
This concludes the proof of the bound~\eqref{eq:growthperturbations} for iteration $t$, finishing Step III.


\paragraph{Step IV: Original and linearized parameters are close:} The difference between the parameter of the original iterate $\vtheta$ and the linearized iterate $\widetilde{\vtheta}$ obey
\begin{align*}
\frac{1}{\stepsize} \norm[2]{\vtheta_\iter - \widetilde{\vtheta}_\iter}
&\leq
\norm[2]{
\sum_{\iteralt=0}^{\iter-1}
\nabla \loss(\vtheta_\iteralt) - \nabla \losslin(\widetilde{\vtheta}_\iteralt)
} \\
&=
\norm[2]{
\sum_{\iteralt=0}^{\iter-1}
\transp{\Jac}(\vtheta_\iteralt) \vr_\iteralt
-
\transp{\mJ} \widetilde{\vr}_\iteralt
} \\
&\leq
\sum_{\iteralt=0}^{\iter-1}
\norm[2]{(\transp{\Jac}(\vtheta_\iteralt) - \transp{\mJ}) \widetilde{\vr}_\iteralt}
+
\norm[2]{\transp{\Jac}(\vtheta_\iteralt) ( \vr_\iteralt -\widetilde{\vr}_\iteralt) } \\
&\mystackrel{(i)}{\le}
\sum_{\iteralt=0}^{\iter-1}
(\epsilon_0 + \epsilon)
\norm[2]{\widetilde{\vr}_\iteralt} 
+
\beta \norm[2]{ \ve_\iteralt } \\
&\mystackrel{(ii)}{\le}
\sum_{\iteralt=0}^{\iter-1}
(\epsilon_0 + \epsilon)
(1-\eta\alpha^2)^\iteralt \norm[2]{\vr_0}
+
2\eta\beta^2  (\epsilon_0 + \epsilon) (1 - \eta \alpha^2)^{\iteralt-1} \iteralt \norm[2]{\vr_0}.
\end{align*}
Here, (i) follows from \eqref{eq:bounddiffjacbs} combined with Assumption \ref{BndSpec} and (ii) follows from~\eqref{eq:growthperturbations} established in step III.
We now proceed by using the formulas for low-order polylogarithms to conclude that
\begin{align*}
\frac{1}{\stepsize} \norm[2]{\vtheta_\iter - \widetilde{\vtheta}_\iter}
&\leq
(\epsilon_0 + \epsilon) \norm[2]{\vr_0}
\left(
\frac{1-(1-\eta\alpha^2)^\tau}{\eta\alpha^2} 
+
2\eta \beta^2 
 \frac{1 - \iter (1-\eta \alpha^2)^{\iter-1}  + (\iter-1) (1-\eta \alpha^2)^{\iter}  }{\eta^2 \alpha^4}
 \right)
  \\
&\leq
(\epsilon_0 + \epsilon) \norm[2]{\vr_0} 
\left(
\frac{1}{\eta\alpha^2} 
+
2 \eta \beta^2 
\frac{1}{\eta^2 \alpha^4} 
\right)
\\
&\leq
(\epsilon_0 + \epsilon) 
\frac{2.5}{\eta\alpha^2} 
\frac{\beta^2}{\alpha^2} \norm[2]{\vr_0} .
\end{align*}
This concludes the proof of~\eqref{eq:propthetakclose} for iteration $t$, completing Step IV. 

\paragraph{Step V: Proof of \eqref{eq:indhypinRBall}:} By the triangle inequality
\begin{align*}
\norm[2]{\vtheta_\iter - \vtheta_0}
&\leq
\norm[2]{\widetilde{\vtheta}_\iter - \vtheta_0}
+
\norm[2]{\vtheta_\iter - \widetilde{\vtheta}_\iter} \\
&\mystackrel{(i)}{\leq}
\norm[2]{\mJ^\dagger\vr_0}
+(\epsilon_0 + \epsilon) \frac{2.5}{\alpha^2} \frac{\beta^2}{\alpha^2} \norm[2]{\vr_0} \\
&\mystackrel{(ii)}{=}
R/2.
\end{align*}
Here, inequality (ii) follows from the definition of $R$ in equation~\eqref{eq:defR}.
Moreover, inequality (i) follows from the bound~\eqref{eq:propthetakclose}, which we just proved, and the fact that, from equation~\eqref{eq:lindifftheta} in Theorem~\ref{lem:rk},
\begin{align*}
\norm[2]{\widetilde{\vtheta}_\iter - \vtheta_0}^2
&=
\sum_{i=1}^n
\innerprod{\vw_i}{\vr_0}^2
\frac{(1 - (1 - \stepsize \sigma^2_i)^\iter)^2}{\sigma_i^2} \\
&\leq
\sum_{i=1}^n
\innerprod{\vw_i}{\vr_0}^2 / \sigma_i^2 \\
&= \norm{\pinv{\mJ} \vr_0 }^2.
\end{align*}
This concludes the proof of \eqref{eq:indhypinRBall} for iteration $t$, completing the proof of Step V and the entire theorem.


\section{Proofs for neural network generators (proof of Theorem~\ref{thm:mainResult})}
\label{pfthm3}


The proof of Theorem~\ref{thm:mainResult} relies on the fact that, in the overparameterized regime, the non-linear least squares problem is well approximated by an associated linearized least-squares problem. Studying the associated linear problem enables us to prove the result.

We apply Theorem \ref{thm:maintechnical},
which ensures that the associated linear problem is a good approximation of the non-linear least squarest problem,
with the non-linear function
\[
f(\mC) = \mA \relu( \mU \mC) \vv
\]
and with the parameter given by $\vtheta=\mC$. Recall that $\vv$ is a fixed vector with half of the entries $1/\sqrt{k}$, and the other half $-1/\sqrt{k}$. 
Let $\Jac(\mC) \in \reals^{m \times nk}$ be the Jacobian of $f$. We have that $\Jac(\mC) = \mA\Jac_G(\mC)$, where $\Jac_G(\mC)$ is the Jacobian of the generator $G$ defined in~\eqref{eq:JacmC}. 
Both $f$ and its Jacobian are random variables because $\mA$ is a random matrix. 
As the reference Jacobian in the associated linear problem, we choose a matrix 
$\mJ = \mA \mJ_G \in \reals^{m \times nk}$
(specified later) that obeys
\[
\mJ \transp{\mJ}
= 
\EX{ \Jac(\mC) \transp{\Jac}(\mC) }
=
\mA 
\underbrace{ \EX{ \Jac_G(\mC) \transp{\Jac}_G(\mC) } }_{\mSigma(\mU)} \transp{\mA}.
\]
Here, expectation is with respect to $\mC$ with iid $\mc N(0,\omega^2)$ parameters, and \emph{not} with respect to $\mA$. 
We apply Theorem \ref{thm:maintechnical} with
\begin{align*}
\alpha =\frac{1}{3\sqrt{2}} \frac{\sqrt{n}}{\sqrt{m}}\sigma_{n}\left(\mSigma(\mU)\right),
\quad \beta = 
2\frac{\sqrt{n}}{\sqrt{m}} \norm{\mU},
\quad
%
\epsilon_0
=2 
\beta
\left( \frac{\log(\frac{2n}{\delta})}{k} \right)^{1/4} ,
\quad
\epsilon = \frac{\xi}{16}\frac{\alpha^4}{\beta^3},
\quad 
\omega = 
\frac{\xi\norm[2]{\vy} }{\beta\sqrt{m}}.
\end{align*}
We next verify that the conditions of Theorem~\ref{thm:maintechnical} are satisfied (specifically, Assumptions \ref{BndSpec}, \ref{assJac2}, \ref{assJac3}) by applying a series of Lemmas. 

Throughout these proofs we use the fact that for a matrix $\mA \in \reals^{m \times n}$ with i.i.d.~$\mathcal{N}(0,\frac{1}{m})$ entries, the bounds
\begin{align*}
\sigma_{\min}(\mA)\ge\frac{\sqrt{n}-(1+\eta)\sqrt{m}}{\sqrt{m}}\quad\text{and}\quad \norm{\mA}\le \frac{\sqrt{n}+(1+\eta)\sqrt{m}}{\sqrt{m}}
\end{align*}
hold with probability at least $1-2e^{-\frac{\eta^2}{2}m}$ which with $\eta=1$ in turn implies that for $m\le \frac{n}{9}$ we have
\begin{align}
\label{eigA}
\sigma_{\min}(\mA)\ge\frac{1}{3}\frac{\sqrt{n}}{\sqrt{m}}\quad\text{and}\quad \norm{\mA}\le 2\frac{\sqrt{n}}{\sqrt{m}}
\end{align}
holds with probability at least $1-2e^{-\frac{m}{2}}$. 
See~\citep[Corollary~5.35]{Vershynin_2012} for a proof of this standard result.

\paragraph{Bound on initial residual:}
We start with bounding the initial residual by applying the following lemma.
\begin{lemma}[{Initial residual~\citep[Lemma~6]{heckel_denoising_2020}}]
\label{lem:InitialResidual}
Consider $G(\mC) = \relu(\mU \mC) \vv$, and let $\mC\in\reals^{n\times k}$ be generated at random with i.i.d.~$\mathcal{N}(0,\omega^2)$ entries. Suppose half of the entries of $\vv$ are $1/\sqrt{k}$ and the other half are $-1/\sqrt{k}$. 
Then, with probability at least $1-\delta$,
\begin{align*}
\norm[2]{G(\mC)} 
\leq \omega \sqrt{8 \log(2n/\delta)}\norm[F]{\mU}.
\end{align*}
\end{lemma}
With this lemma in place, the initial residual can be upper bounded as follows
\begin{align}
\norm[2]{\vr_0}
&\leq 
\norm[2]{\vy} + \norm[2]{\mA G(\mC_0)} \nonumber \\
&\mystackrel{(i)}{\leq} 
\norm[2]{\vy} + 2\norm[2]{G(\mC_0)} \nonumber \\
&\mystackrel{(ii)}{\leq}
3 \norm[2]{\vy}.
\label{eq:boundinitres}
\end{align}
Here (i) holds with probability at least $1-e^{-\frac{m}{2}}$ using the fact that $\mA$ has i.i.d.~Gaussian entries that are independent of $G(\mC_0)$, and 
for (ii) we used that, by Lemma~\ref{lem:InitialResidual},
\begin{align}
\label{eq:boundinitgen}
\norm[2]{G(\mC_0)}
&\leq  \omega \sqrt{8 \log(2n/\delta)}\norm[F]{\mU}\nonumber\\
&\leq  \omega \sqrt{8 \log(2n/\delta)} \sqrt{n}\norm{\mU}\nonumber\\
&\mystackrel{(i)}{=} \xi \sqrt{2 \log(2n/\delta)} \norm[2]{\vy}\nonumber\\
&\mystackrel{(ii)}{\leq} \norm[2]{\vy},
\end{align}
where (i) follows from $\omega = \frac{\xi\norm[2]{\vy} }{\beta\sqrt{m}} = \frac{\xi\norm[2]{\vy} }{2 \sqrt{n} \norm{\mU}}$ and for (ii) we used the fact that $\xi\le \frac{1}{\sqrt{2 \log(2n/\delta)}}$.

%
\paragraph{Verifying Assumption~\ref{BndSpec}:} Note that
\begin{align*}
\sigma_{\min}\left(\mJ\right)=\sigma_{\min}\left(\mA\mJ_G\right)\ge \sigma_{\min}(\mA)\sigma_{\min}(\mJ_G)\ge \frac{1}{3}\frac{\sqrt{n}}{\sqrt{m}}\sigma_{n}\left(\mSigma(\mU)\right)\ge \alpha\sqrt{2}.
\end{align*}
We next show that the norm of the reference Jacobian and the Jacobian are bounded, with the lemma below.

\begin{lemma}[Spectral norm of Jacobian
{\citep[Lemma~5]{heckel_denoising_2020}}]
\label{lem:boundJac}
Consider $G(\mC) = \relu(\mU \mC) \vv$ with $\vv\in \reals^k$ and $\mU \in \reals^{n\times k}$ and associated Jacobian $\Jac_G(\mC)$~\eqref{eq:JacmC}, and let $\mJ_G$ be any matrix obeying 
$\mJ_G \transp{\mJ}_G = \EX{\Jac_G(\mC) \transp{\Jac}_G(\mC)}$, 
where the expectation is over a matrix $\mC$ with iid $\mc N(0,\omega^2)$ entries.
Then
\[
\norm{\Jac_G(\mC)}
\leq
\norm[2]{\vv}\norm{\mU}
\quad
\text{and}
\quad
\norm{\mJ_G}
\leq
\norm[2]{\vv}
\norm{\mU}.
\]
\end{lemma}
By Lemma~\ref{lem:boundJac}, with $\norm[2]{\vv}=1$,
\[
\norm{\mJ} 
=
\norm{\mA \mJ_G}
\leq 
\norm{\mA} \norm{\mJ_G}
\leq 2 \sqrt{n/m} \norm{\mU} = \beta,
\]
where the last inequality follows from Lemma~\ref{lem:boundJac}, with $\norm[2]{\vv}=1$, and by using that, with high probability, $\norm{\mA} \leq 2 \sqrt{n/m}$ per \eqref{eigA}. 
Analogously, we obtain $\norm{\Jac(\mC)} \leq \beta$, for all $\mC$, with high probability. This completes the verification of Assumption \ref{BndSpec}. 
%

\paragraph{Verifying Assumption \ref{assJac2}:} 
To verify the assumption, we first state a concentration lemma from~\cite{heckel_denoising_2020}.
\begin{lemma}[Concentration lemma {\citep[Lemma~3]{heckel_denoising_2020}}]
\label{lem:ConcentrationLemma}
Consider $G(\mC) = \relu(\mU \mC) \vv$ with $\vv\in \reals^k$ and $\mU \in \reals^{n\times k}$ and associated Jacobian $\Jac_G(\mC)$~\eqref{eq:JacmC}. 
Let $\mC\in\reals^{n\times k}$ be generated at random with i.i.d.~$\mathcal{N}(0,\omega^2)$ entries. Then, with probability at least $1-\delta$,
\begin{align*}
\norm{ \Jac_G(\mC)\Jac^T_G(\mC) - \mSigma(\mU) }
\le\norm{\mU}^2 \sqrt{ \log\left(\frac{2n}{\delta}\right)  \sum_{\ell=1}^k v_\ell^4 }.
\end{align*}
\end{lemma}
Using the fact that $\sum_\ell^k v_\ell^4 = \frac{1}{k}$ by Lemma~\ref{lem:ConcentrationLemma} we have 
\begin{align}
\label{eq:fromconcentration}
\norm{\Jac_G(\mC_0)\Jac^T_G(\mC_0)-\mSigma(\mU)}
\leq 
\norm{\mU}^2 \sqrt{\frac{\log\left(2n/\delta\right)}{k}}.
\end{align}
To show that \eqref{eq:fromconcentration} implies the condition in~\eqref{eq:assJacJclose}, we use the following lemma.
\begin{lemma}[{\citep[Lem.~6.4]{oymak_generalization_2019}}]
\label{eq:lemmaC1impC2}
Let  $\mX \in \reals^{n\times \N}$, $\N \geq n$ and let $\mSigma$ be $n \times n$ psd matrix obeying $\norm{ \mX \transp{\mX} - \mB } \leq \tilde \epsilon^2$, for a scalar $\tilde \epsilon \geq 0$. Then there exists a matrix $\mJ_G \in \reals^{n \times \N}$ obeying $\mSigma = \mJ_G \transp{\mJ}_G$ such that
\[
\norm{\mJ_G - \mX} \leq \tilde 2\epsilon.
\]
\end{lemma}

From Lemma~\ref{eq:lemmaC1impC2} combined with equation~\eqref{eq:fromconcentration}, we have that there exists a matrix $\mJ_G \in \reals^{n\times N}$ that obeys 
\begin{align*}
\norm{ \mJ_G - \Jac_G(\mC_0)} 
\leq 
2 \norm{\mU} \left( \frac{\log( 2n / \delta )}{k} \right)^{1/4}.
\end{align*}
Using this inequality, as well as that 
$\norm{\mA}\le 2\frac{\sqrt{n}}{\sqrt{m}}$, per~\eqref{eigA}, we get
\begin{align*}
\norm{ \mJ - \Jac(\mC_0)}
&=
\norm{\mA ( \mJ_G - \Jac_G(\mC_0) )}\\
&\leq
\norm{\mA}
\norm{ \mJ_G - \Jac_G(\mC_0)} \\
&\leq
2 \frac{\sqrt{n}}{\sqrt{m}}
2 \norm{\mU} \left( \frac{\log( 2n / \delta )}{k} \right)^{1/4} \\
&\leq
2 \beta \left( \frac{\log( 2n / \delta )}{k} \right)^{1/4}\\
&= \epsilon_0,
\end{align*}
as desired. This concludes the proof of Assumption~\ref{assJac2}. 

This part of the proof also specifies our choice of the reference Jacobian $\mJ = \mA \mJ_G$ as a matrix that is $\epsilon_0$ close to the Jacobian at initialization, $\Jac(\mC_0)$, and that exists by Lemma~\ref{eq:lemmaC1impC2} above.

\paragraph{Verifying Assumption \ref{assJac3}:} 
Verification of the assumption requires us to control the perturbation of the Jacobian matrix around a random initialization. We begin with the following lemma from \cite{heckel_denoising_2020}.
\begin{lemma}[Jacobian perturbation around initialization {\citep[Lemma~7]{heckel_denoising_2020}}]
\label{lem:jacobianperturbation}
Let $\mC_0$ be a matrix with i.i.d. $\mathcal \mathcal{N}(0,\omega^2)$ entries.
Then, for all $\mC$ obeying
\begin{align*}
 \norm{\mC - \mC_0} \le \omega \widetilde{R}\quad\text{with}\quad \widetilde{R} \leq \frac{1}{2}\sqrt{k},
 \end{align*} 
the Jacobian mapping~\eqref{eq:JacmC}associated with the generator $G(\mC) = \relu(\mU \mC) \vv$ obeys
\begin{align*}
\norm{\Jac_G(\mC) - \Jac_G(\mC_0)}
\leq
\norm[\infty]{\vv}
 2 (k \widetilde{R})^{1/3}\norm{\mU},
\end{align*}
with probability at least  $1- n e^{- \frac{1}{2} \widetilde{R}^{4/3} k^{7/3} }$.
\end{lemma}


In order to verify Assumption \ref{assJac3}, 
first note that the radius in the theorem, defined in equation~\eqref{eq:defR}, obeys
\begin{align*}
R 
&=2\norm[2]{ \mJ^\dagger\vr_0}
+5\frac{\beta^2}{\alpha^4}(\epsilon_0 + \epsilon)   \norm[2]{\vr_0}\\
&\mystackrel{(i)}{\le}
\left(\frac{\sqrt{2}}{\alpha}+
\frac{5}{8\beta}
\right) \norm[2]{\vr_0} \\
&\mystackrel{(ii)}{\le}
9 \frac{1}{\alpha} \norm[2]{\vy} \\
&\mystackrel{(iii)}{=} 9 \omega\frac{\sqrt{m}}{\xi} \frac{\beta}{\alpha}\\
&\mystackrel{(iv)}{\le} \omega\frac{1}{(4\cdot 16)^3}\xi^3\frac{\alpha^{12}}{\beta^{12}}\sqrt{k} \\
&:=\omega \widetilde{R}.
\end{align*}
Here, 
(i) follows from the fact that $\norm[2]{\mJ^\dagger\vr_0}\le \frac{1}{\alpha\sqrt{2}}\norm[2]{\vr_0}$, 
and using that 
$\epsilon_0 + \epsilon \leq 2 \epsilon = \frac{1}{8} \xi \frac{\alpha^4}{\beta^3} \leq \frac{1}{8} \xi \frac{\alpha^4}{\beta^3}$
(ii) from $\beta \geq \alpha$ and from the bound on the initial residual~\eqref{eq:boundinitres},
(iii) from $\omega=\frac{\xi\norm{\vy}}{\beta\sqrt{m}}$
and finally (iv) follows from the assumption \eqref{eq:thmoverparamass} which is equivalent to 
\[
k
\geq
m \xi^{-8}9^2 64^6 \left(\frac{\beta}{\alpha}\right)^{26}   
=
m \xi^{-8}9^2 64^6 \left( 6 \sqrt{2} \frac{\norm{\mU}^2}{\sigma_n^2} \right)^{26}
= C^2 \frac{\kappa_\vu^{26}}{\xi^8}m. 
\]
For this choice of radius by Lemma~\ref{lem:jacobianperturbation} 
and by using 
$\norm{\mA}\le 2\frac{\sqrt{n}}{\sqrt{m}}$ (per~\eqref{eigA}) we have
\begin{align*}
\norm{\mA\Jac_G(\mC) - \mA\Jac_G(\mC_0)}
\le&\norm{\mA}\norm{\Jac_G(\mC) - \Jac_G(\mC_0)}\\
\leq&2\frac{\sqrt{n}}{\sqrt{m}}
2\norm[\infty]{\vv}
  (k \widetilde{R})^{1/3}\norm{\mU}\\
  =&2\beta\frac{1}{\sqrt{k}}
  (k \widetilde{R})^{1/3}\\
=&\frac{1}{32}\xi\frac{\alpha^4}{\beta^3}\\
=&\frac{\epsilon}{2}
\end{align*}
holds with probability at least
\begin{align*}
1-n e^{- \frac{1}{2} \widetilde{R}^{4/3} k^{7/3} }\mystackrel{(i)}{\ge} 1-ne^{-k^2}
\end{align*}
where in (i) we used \eqref{eq:thmoverparamass}. Therefore, Assumption~\ref{assJac3} holds with high probability by our choice of $\epsilon =\frac{\xi}{30}\frac{\alpha^4}{\beta^3}$.

\paragraph{Concluding the proof of Theorem~\ref{thm:mainResult}:}
To begin, let $\vc^\ast$ be a solution to the optimization problem
\begin{align*}
\vc^\ast 
\in
\underset{\vc}{\arg\min}\quad \frac{1}{2}\norm[2]{\mJ_G\vc - \vx^\ast}^2.
\end{align*}
To complete the proof of Theorem~\ref{thm:mainResult} let us consider the linearized optimization problem which takes the form
\begin{align*}
\min_{\vc}
\losslin(\vc)
=
\frac{1}{2}
\norm[2]{ \mA G(\vc_0) + \mA \mJ_G (\vc - \vc_0) - \vy}^2,
\end{align*}
with corresponding iterates given by
\begin{align*}
\widetilde{\vc}_{\iter+1}=\widetilde{\vc}_\iter-\eta\nabla \losslin
(\widetilde{\vc}_\iter).
\end{align*}
Here, $\vc$ is the vectorized version of $\mC$, with a slight abuse of notation.
With this notation, we conclude the proof as
\begin{align*}
\norm[2]{G(\mC_\infty)-\vx^{\ast}}
&\le
\norm[2]{G(\mC_\infty)-\mJ_G\widetilde{\vc}_\infty}+\norm[2]{\mJ_G\widetilde{\vc}_\infty-\vx^\ast} \\
&\le
\xi \norm[2]{\vx^\ast}
+
C
\left( \sum_{i=1}^n \frac{1}{\sigma_i^2} \innerprod{\vw_i}{\vx^\ast}^2 \right)
\sum_{i > 2m/3} \sigma_i^2,
\end{align*}
where we used the bounds
\begin{align}
\label{eq:detailsforgmcinw}
\norm[2]{G(\mC_\infty)-\mJ_G\widetilde{\vc}_\infty}
\leq
\xi \norm[2]{\vx^\ast} 
\end{align}
and
\begin{align}
\label{eq:fromlincase}
\norm[2]{\mJ_G\widetilde{\vc}_\infty-\vx^\ast}
\leq
C \left( \sum_{i=1}^n \frac{1}{\sigma_i^2} \innerprod{\vw_i}{\vx^\ast}^2 \right)
\sum_{i > 2m/3} \sigma_i^2.
\end{align}
The bound~\eqref{eq:fromlincase} follows from Theorem~\ref{prop:specialcase} by noting that 
$\vw_1,\ldots,\vw_n$ are the left singular vectors of $\mJ_G$ with associated singular values $\sigma_1 \geq \ldots \geq \sigma_n$ (because 
$\mJ_G \transp{\mJ}_G = \mSigma(\mU))$. 

It remains to prove the bound~\eqref{eq:detailsforgmcinw}.
With $\Jac_G(\va,\vb) = \int_0^1 \Jac_G(s \vb - (1-s) \va) ds$, at $t= + \infty$,
\begin{align*}
\norm[2]{G(\mC_\infty)-\mJ_G\widetilde{\vc}_\infty}
&=\norm[2]{\Jac_G(\mC_\infty,\mtx{0})
\text{vect}(\mC_\infty)-\mJ_G\widetilde{\vc}_\infty} \\
%
%
&\leq 
\norm[2]{\Jac_G(\mC_\infty,\mtx{0}) (\text{vect}(\mC_\infty) -\widetilde{\vc}_\infty)}
+
\norm[2]{\Jac_G(\mC_\infty,\mtx{0})\widetilde{\vc}_\infty-\mJ_G\widetilde{\vc}_\infty} \\
&\leq
\norm{\Jac_G(\mC_\infty,\mtx{0})}
\norm[2]{ \text{vect}(\mC_\infty) - \widetilde{\vc}_\infty}
+
\norm[2]{\Jac_G(\mC_\infty,\mtx{0})-\mJ_G}
\norm[2]{\widetilde{\vc}_\infty} \\
&\mystackrel{(i)}{\le}
\frac{\sqrt{m}}{\sqrt{n}}\frac{\beta}{2}
\norm[2]{ \text{vect}(\mC_\infty) - \widetilde{\vc}_\infty }
+\frac{\sqrt{m}}{2\sqrt{n}}(\epsilon+\epsilon_0)\norm[2]{\widetilde{\vc}_\infty} \\
&\mystackrel{(ii)}{\le}
\frac{\beta}{2}
\norm[2]{ \text{vect}(\mC_\infty) - \widetilde{\vc}_\infty }
+
\frac{1}{2} (\epsilon+\epsilon_0) \frac{1}{\alpha} \norm[2]{\vx^\ast}.
\end{align*}
In the above 
(i) follows from $\norm{\mJ_G(\mC)} \leq \norm{\mU} = \frac{\sqrt{m}}{2\sqrt{n}} \beta$ 
(recall that $\beta = 
2\frac{\sqrt{n}}{\sqrt{m}} \norm{\mU}$)
and from the bound
\[
\norm[2]{\Jac_G(\mC_\infty,\mtx{0})-\mJ_G}
\leq
\norm[2]{\Jac_G(\mC_\infty,\mtx{0}) - \Jac_G(\mC_0)}
+
\norm[2]{\Jac_G(\mC_0) - \mJ_G}
\leq
\frac{\sqrt{m}}{2\sqrt{m}} (\epsilon_0 + \epsilon).
\]
Moreover, (ii) follows from $m\leq n$ and 
$\norm[2]{\vc_\infty} \leq \norm[2]{\vc^\ast} \le \frac{\norm[2]{\vx^\ast}}{\sigma_{\min}(\mJ_G)}$. 
%
We can now apply Theorem~\ref{thm:maintechnical} equation \eqref{eq:propthetakclose} to bound the first term on the right-hand-side above to obtain
\begin{align*}
\norm[2]{G(\mC_\infty)-\mJ_G\widetilde{\vc}_\infty}
&\le 1.25 \frac{\beta^3}{\alpha^4} (\epsilon_0 + \epsilon)\norm[2]{\vr_0}+\frac{1}{2\alpha}(\epsilon+\epsilon_0)\norm[2]{\vx^\ast}\\
&\mystackrel{(i)}{\leq}
 7.5 \frac{\beta^3}{\alpha^4} (\epsilon_0 + \epsilon)\norm[2]{\vx^\ast}+\frac{1}{2\alpha}(\epsilon+\epsilon_0)\norm[2]{\vx^\ast}\\
&\mystackrel{(ii)}{\leq}
16 \frac{\beta^3}{\alpha^4} \epsilon\norm[2]{\vx^\ast}\\
&\mystackrel{(iii)}{=} \xi\norm[2]{\vx^\ast}.
\end{align*}
Here, (i) follows from $\norm[2]{\vr_0} \leq 3 \norm[2]{\vy} = 3 \norm[2]{\mA \vx^\ast} \leq 6 \norm{\vx^\ast}$, where we used~\eqref{eq:boundinitres} combined with the fact that $\norm[2]{\mA\vx^\ast}\le 2\norm[2]{\vx^\ast}$.
Moreover, (ii) follows from $\frac{\beta}{\alpha}\geq 1$ and $\epsilon_0\leq \epsilon$ 
and finally (iii) from the choice 
$\epsilon=\frac{\xi}{16}\frac{\alpha^4}{\beta^3}$. 
This concludes the proof of the bound~\eqref{eq:detailsforgmcinw} and the proof of the theorem.

\end{document}